\documentclass[sigconf,nonacm]{acmart}
\renewcommand\footnotetextcopyrightpermission[1]{}
\usepackage{amsmath}
\usepackage{tabularx}
\usepackage{subcaption}
\usepackage{fvextra}
\fvset{breaklines=true,breakanywhere=true,fontsize=\footnotesize,%
  breaksymbolleft=\textcolor{black!45}{\tiny\ensuremath{\hookrightarrow}}}
\makeatletter
\AtBeginEnvironment{Verbatim}{\let\@vspace\@vspace@orig
  \let\@vspacer\@vspacer@orig}
\AtBeginEnvironment{Verbatim*}{\let\@vspace\@vspace@orig
  \let\@vspacer\@vspacer@orig}
\makeatother
\makeatletter
\newcommand{\rawvspace}[1]{%
  \begingroup\let\@vspace\@vspace@orig\let\@vspacer\@vspacer@orig\vspace{#1}\endgroup}
\makeatother

\newcommand{\cmark}{\ding{51}}
\newcommand{\xmark}{\ding{55}}

\begin{document}

\title{\textsc{TRACE-TS}: Attribution-Grounded and Traceable Sensor-Language Reasoning for Human Activity Understanding}

% acmart flags \vspace outside floats; the manual topmatter below is a
% deliberate preprint layout, so restore the original \vspace there.
\makeatletter
\let\@vspace@save\@vspace \let\@vspacer@save\@vspacer
\let\@vspace\@vspace@orig \let\@vspacer\@vspacer@orig
\makeatother
\twocolumn[
\begin{@twocolumnfalse}

% ---------- TITLE ----------
\begin{center}

{\LARGE\bfseries
TRACE-TS: Attribution-Grounded and Traceable Sensor-Language Reasoning for Human Activity Understanding
\par}

\vspace{1.2em}

{\large
Sparsh Rastogi$^{1,2,*}$ \quad
Tanmay Kumar$^{1,*}$ \quad
Baiyu Chen$^{1,*,\dagger}$ \quad
Jatin Bedi$^{2}$ \quad
Zechen Li$^{1,\dagger}$ \quad
Flora D. Salim$^{1,\dagger}$
\par}

\addvspace{0.8em}

{\normalsize
$^{1}$University of New South Wales, Sydney, Australia\\
$^{2}$Thapar Institute of Engineering and Technology, Patiala, India
\par}

\addvspace{0.8em}

{\small
\texttt{\{tanmay.kumar,breeze.chen,zechen.li,flora.salim\}@unsw.edu.au}\\
\texttt{\{srastogi\_be22,jatin.bedi\}@thapar.edu}
\par}

\vspace{0.5em}

{\small
$^{*}$Equal contribution \hspace{2em}
$^{\dagger}$Corresponding authors
}

\end{center}

\vspace{1em}

\end{@twocolumnfalse}
]
\makeatletter\let\@vspace\@vspace@save \let\@vspacer\@vspacer@save\makeatother

\noindent{\bfseries\large Abstract\par}
\addvspace{0.4em}
\noindent
Wearable sensors capture fine-grained motion patterns that support rich behavioral understanding, yet most existing methods reduce these signals to activity labels. Recent LM-based approaches generate natural-language explanations for sensor data, but their reasoning is weakly grounded in the underlying signal, leading to fluent yet unverifiable explanations. We introduce \textsc{TRACE-TS} (Traceable Reasoning with Attribution-Grounded Evidence), a framework for structured and signal-grounded reasoning over wearable time series. \textsc{TRACE-TS} uses attribution from an expert classifier to identify salient spatio-temporal sensor regions, uses them to construct DAG reasoning traces with explicit evidence provenance, and trains a compact language model to generate these traces through gated cross-attention over sensor memory tokens. At inference, the adapted model jointly outputs the activity prediction and its reasoning trace, without requiring attribution computation or teacher guidance. We introduce \emph{Semantic Node Match} (SNM), an LLM-as-judge metric that diagnoses reasoning fidelity at the observation, inference, and synthesis levels, localizing hallucinated observations and broken evidence chains missed by standard NLG metrics. Across seven wearable benchmarks, \textsc{TRACE-TS} achieves the best average accuracy and F1 among all evaluated methods (\textbf{84.43\%}/\textbf{81.24\%}), and outperforms the best LLM-based baseline by \textbf{17.96}\% in F1. Our code is available at \url{https://github.com/SparshRastogi/TRACE-TS}.

\addvspace{0.8em}

\thispagestyle{plain}
\pagestyle{plain}

\section{Introduction}
\label{sec:intro}

Wearable sensors capture fine-grained motion patterns that reflect behavior, health, and intent \cite{biswas2025wearable, smets2018large, large-scale-training}. As people walk, exercise, rest, or perform daily activities, their body movements produce temporal signals that encode rich behavioral cues beyond discrete activity labels~\cite{stanfordheartcount, covid}. With the widespread adoption of mobile and wearable devices, including phones, watches, and emerging XR systems, these signals can now be collected continuously and at scale in real-world settings \cite{truslow2024understanding, allofus}. However, most wearable time-series (TS) models remain optimized for label prediction, mapping each window to a fixed activity class. This classification-centric paradigm typically exposes only the final label, leaving the supporting sensor evidence, intermediate behavioral cues, and reasoning structure implicit or available only through post-hoc analysis \cite{bian2026foundation}.

Recent work has explored language models (LMs) to move beyond label prediction, either by prompting LMs with serialized sensor windows or by aligning sensor representations with LMs through instruction tuning, sensor-language supervision, and time-series LM adaptation~\cite{hargpt, llasa, langer2025opentslm,li2025sensorllm}. While some of these methods enable more expressive outputs than fixed labels, their reasoning remains weakly grounded in the underlying signal: explanations may cite uninformative channels, describe absent temporal patterns, or rely on unsupported claims. As a result, the output can appear coherent while failing to reflect the recorded sensor evidence.
\begin{figure}
    \centering
    \includegraphics[width=\linewidth]{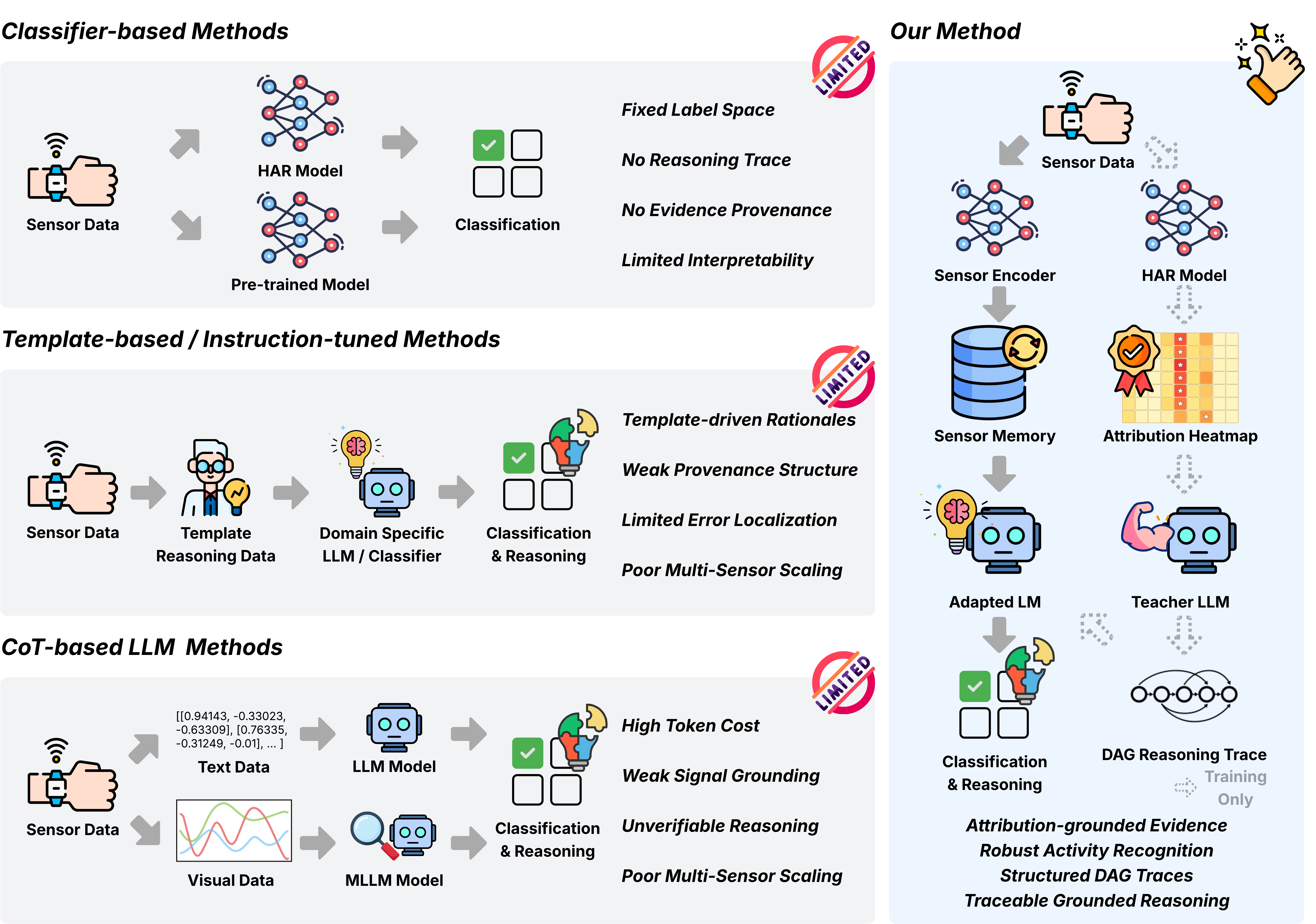}
    \Description{Comparison of classifier-based, template-based or instruction-tuned, and LLM-based method families with the proposed TRACE-TS approach, which adds attribution-grounded evidence and structured DAG reasoning traces.}
    \caption{Representative Method Families for human activity understanding.}
    \label{fig:intro}
\end{figure}

As illustrated in Figure~\ref{fig:intro}, this gap highlights a need for sensor-language reasoning methods that go beyond explanations. 
First, explanations should be \textbf{grounded in sensor evidence}: they should remain connected to the sensor signal rather than relying on language priors or activity-label semantics.
Otherwise, a model may produce plausible descriptions that refer to patterns absent from the signal or temporal changes unsupported by the recorded data.
Second, explanations should be \textbf{traceable}: the model should expose how low-level observations support intermediate inferences and final activity predictions, instead of producing unstructured free-form rationales.
Third, evaluation should be \textbf{diagnostic}: it should assess whether explanations are consistent with the sensor-derived reasoning process, since standard NLG metrics such as ROUGE, METEOR, and BERTScore mainly measure surface similarity and provide limited support for diagnosing unsupported claims or inconsistent reasoning steps~\cite{sivalingam2026llm, arai-etal-2025-evaluating, zhou2025enhancing}.

To address these requirements, we introduce \textsc{TRACE-TS} (Traceable Reasoning with Attribution-Grounded Evidence), a framework for generating structured reasoning traces over wearable sensor time series. 
\textsc{TRACE-TS} first uses attribution from an expert classifier to identify salient spatio-temporal sensor regions, which serve as evidence for a teacher large language model (LLM) to construct directed acyclic graph (DAG) reasoning traces with explicit provenance links. 
A compact student LM is then trained to generate these traces by attending to sensor memory tokens through gated cross-attention adapters. 
At inference, the student directly consumes raw sensor inputs and jointly outputs the activity label and its reasoning trace, without requiring the expert classifier, attribution methods, or teacher LLM.
To diagnose reasoning quality, we introduce Semantic Node Match (SNM), which compares predicted and reference traces at the observation, inference, and synthesis levels, enabling localized analysis of hallucinated observations and broken evidence chains. Concretely, this work contributes:

\begin{itemize}

\item \textbf{Attribution-grounded Reasoning Framework.}
We introduce \textsc{TRACE-TS}, a framework that uses attribution from an expert HAR classifier to construct evidence-grounded reasoning supervision, enabling a compact LM to generate activity predictions and structured reasoning traces from wearable sensor inputs.

\item \textbf{Structured and Traceable Reasoning.}
We propose a DAG-based reasoning format with \texttt{based\_on} provenance edges, linking sensor observations, intermediate inferences, synthesis conclusions, and activity predictions for node-level inspection and error localization.

\item \textbf{Structure-aware Reasoning Evaluation.}
We introduce \emph{Semantic Node Match}, an LLM-as-judge framework comparing predicted and reference traces across observation, inference, and synthesis levels, capturing fine-grained reasoning errors missed by standard NLG metrics.

\end{itemize}

\begin{figure*}
    \centering
    \includegraphics[width=\linewidth]{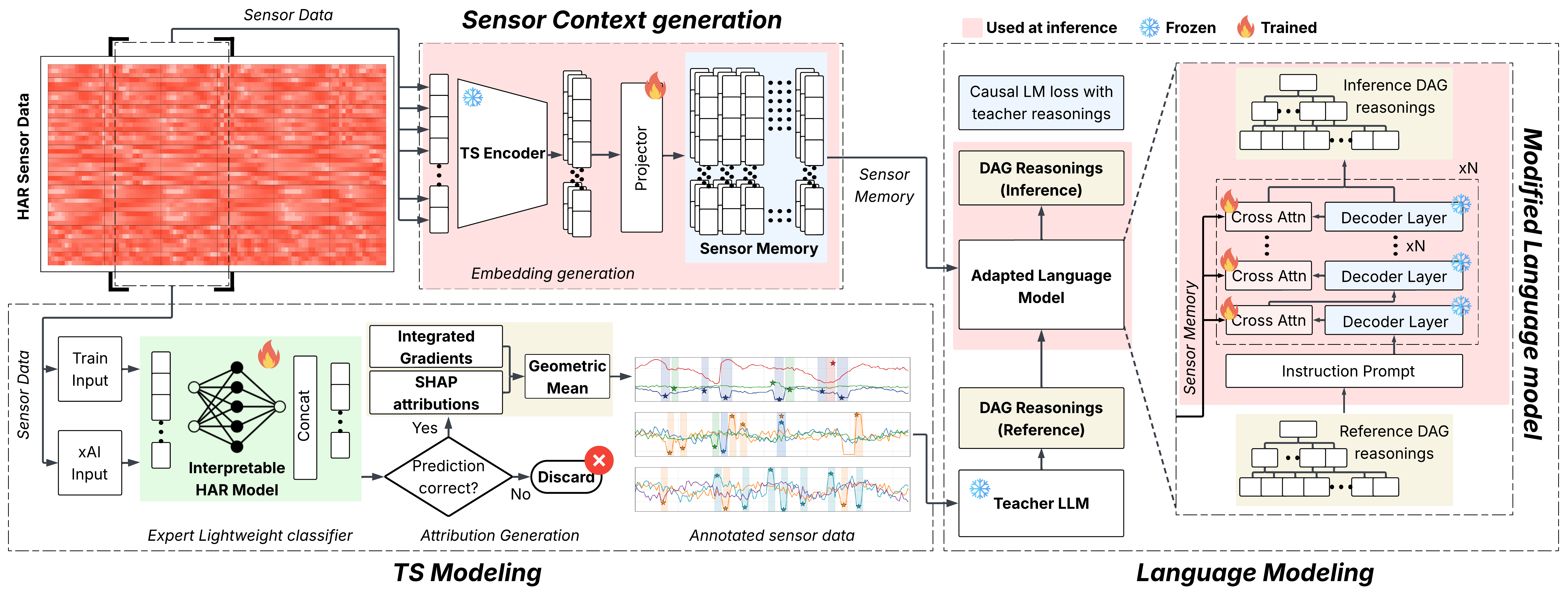}
    \Description{Pipeline diagram of the TRACE-TS framework: sensor data is processed by an interpretable HAR classifier with IG and SHAP attribution, a teacher LLM generates reference DAG reasonings, and a frozen student language model with gated cross-attention adapters learns to generate reasoning from sensor memory tokens.}
    \caption{An illustrative representation of the overall methodology of the proposed TRACE-TS framework.}
    \label{fig:overview}
\end{figure*}

\section{Related Work}

\paragraph{Sensor Representation Learning.}
Early wearable HAR methods mainly formulate sensor understanding as a supervised classification task, using convolutional \cite{moya2018convolutional, ismail2020inceptiontime, andrade2022human}, recurrent \cite{ordonez2016deep, guan2017ensembles, zhao2018deep}, and attention-based architectures \cite{murahari2018attention, abedin2021attend, dirgova2022wearable}. These models achieve strong benchmark performance but are often tied to fixed label spaces, participant distributions, and device configurations, limiting transfer to more diverse real-world settings \cite{chang2020systematic, cai2025towards, dhekane2025transfer}. More recent work improves robustness through large-scale pretraining and self-supervised objectives \cite{yuan2024self, deldari2024crossl}, including masked reconstruction \cite{haresamudram2020masked, xu2025lsm, zhang2026mopformer, muhammad2026cgmjepalearningconsistentcontinuous, li2026glucofm}, contrastive learning \cite{deldari2022cocoa}, synthetic data generation, and cross-dataset adaptation \cite{zhang2024unimts, leng2023generating, miao2026wonderwall, wei2025one, hong2024crosshar, chen2026anymo}. However, these methods still primarily optimize for classification or representation transfer, offering limited support for explaining how specific sensor patterns support a prediction \cite{bian2026foundation}.

\paragraph{LLM-based Reasoning over Sensor Time-series.}
Recent work has explored LLMs for wearable time-series, aiming to generate natural-language explanations beyond discrete activity labels \cite{10.24963/ijcai.2024/921, Liu_2025}. One line of work prompts LLMs with serialized or summarized sensor signals, sometimes using chain-of-thought prompting to elicit reasoning \cite{xue2023promptcast}. While simple, such approaches rely on inefficient text representations of continuous signals and can produce explanations that are weakly grounded in the actual sensor window. Another line aligns sensor representations with LLMs through instruction tuning, shared embedding spaces, multimodal supervision, or cross-attention fusion \cite{li2025sensorllm, llasa, langer2025opentslm, chen2025comodo, jin2024timellmtimeseriesforecasting}. These methods improve sensor-language coupling, but their explanations are typically generated as free-form text, without explicit provenance linking observations to inferences. Agentic approaches further incorporate retrieval or external memory to improve reasoning \cite{li2026zara,chen2026efficientevidencegroundedmobilityprediction}, but often require multi-step inference and are less suitable for efficient deployment. In contrast, \textsc{TRACE-TS} constructs attribution-grounded DAG reasoning traces during training and distills them into a compact sensor-conditioned LM, enabling single-pass activity prediction and reasoning trace generation with structure-aware evaluation.

\section{Methodology}
\label{sec:method}

\textsc{TRACE-TS} learns to generate structured reasoning traces through a two-stage framework, as shown in Figure~\ref{fig:overview}. 
The first stage constructs reasoning supervision, where an expert HAR classifier identifies prediction-relevant regions using attribution, and a teacher LLM converts these regions into DAG traces with evidence links. 
The second stage distills this supervision into a compact sensor-conditioned language model. 
A frozen TS encoder maps the input signal into sensor memory tokens, which are injected into a frozen student LM through gated cross-attention adapters. 
At inference, only the components shaded light red in Figure~\ref{fig:overview} are used: \textsc{TRACE-TS} takes a raw sensor window and generates both the activity prediction and corresponding reasoning trace, without running the expert classifier, attribution methods, or teacher LLM.

\paragraph{Problem Formulation.} Let $\mathbf{X}\in\mathbb{R}^{T\times C}$ denote a multivariate sensor window with $T$ timesteps and $C$ channels, and let $\mathcal{Y}$ be the activity label set. 
Our goal is to learn a model $f_\theta$ that maps $\mathbf{X}$ to an activity prediction $y\in\mathcal{Y}$ and a structured reasoning trace $\mathcal{G}=(\mathcal{V},\mathcal{E})$.
We represent $\mathcal{G}$ as a \textbf{directed acyclic graph} (DAG). 
Its nodes operate at different levels: \textit{observation nodes} $O_i\in\mathcal{V}_{\mathrm{obs}}$ describe localized signal patterns, \textit{inference nodes} $I_j\in\mathcal{V}_{\mathrm{inf}}$ combine observations into higher-level evidence, a \textit{synthesis node} $S$ summarizes the inferred evidence, and an \textit{activity node} $v_{\mathrm{act}}$ gives the predicted label. 
Edges in $\mathcal{E}$ encode \texttt{based\_on} relations, indicating which lower-level nodes support each higher-level claim.
Each observation node is associated with a temporal interval $[t_i^s,t_i^e]$ and a subset of channels $\mathcal{C}_i\subseteq[C]$,  linking each claim to a sensor subregion $O_i\mapsto\mathbf{X}_{t_i^s:t_i^e,\mathcal{C}_i}$, where $1\leq t_i^s<t_i^e\leq T$.
We use an attribution map $\mathbf{A}\in\mathbb{R}^{T\times C}$ from an expert classifier to identify salient regions.
An observation region is salient when its average attribution score exceeds a threshold $\delta$:
$
\frac{1}{|\mathcal{C}_i|(t_i^e - t_i^s + 1)}
\sum_{c \in \mathcal{C}_i} \sum_{t=t_i^s}^{t_i^e} \mathbf{A}_{t,c}
\;\geq\; \delta .
$
These regions provide evidence for traces.

\subsection{Stage 1: Reasoning Distillation}
The first stage constructs structured reasoning supervision from attribution-supported sensor evidence. Given a training sample $\mathbf{X}$ and label $y$, we first obtain salient spatio-temporal regions from an expert HAR classifier, serialize these regions into a closed evidence format, and then prompt a teacher LLM $\mathcal{T}$ to generate a structured DAG reasoning trace that explicitly links each reasoning step to its supporting evidence, providing a traceable reasoning process that enables provenance tracking, fine-grained error localization, and verification of intermediate reasoning steps.

\paragraph{Expert Classifier and Attribution Extraction.}
We train an HAR classifier $\mathcal{C}$ on the train set and compute a per-timestep, per-channel attribution map $\mathbf{A}\in\mathbb{R}^{T\times C}$. 
To reduce dependence on a single attribution method, we combine Integrated Gradients (IG) \cite{sundararajan2017axiomatic} with SHAP.
IG assigns importance to each feature $(t,c)$ by integrating gradients along the path from a baseline $\mathbf{X}'$ to the input $\mathbf{X}$:
\begin{equation}
\begin{aligned}
\mathbf{A}^{\mathrm{IG}}_{t,c}
&= (X_{t,c} - X'_{t,c}) \\
&\quad \cdot \int_0^1 
\frac{\partial\, \mathcal{C}(\mathbf{X}' + \alpha(\mathbf{X} - \mathbf{X}'))}
{\partial X_{t,c}}\, d\alpha .
\end{aligned}
\label{eq:ig}
\end{equation}
SHAP provides a complementary Shapley-value attribution estimate $\mathbf{A}^{\mathrm{SHAP}}$ \cite{lundberg2017unified}. 
We normalize both attribution maps to be non-negative and fuse them by geometric mean:
\begin{equation}
\mathbf{A}_{t,c}
=
\sqrt{
\tilde{\mathbf{A}}^{\mathrm{IG}}_{t,c}
\cdot
\tilde{\mathbf{A}}^{\mathrm{SHAP}}_{t,c}
},
\label{eq:attr}
\end{equation}
where $\tilde{\mathbf{A}}^{\mathrm{IG}}$ and $\tilde{\mathbf{A}}^{\mathrm{SHAP}}$ denote normalized attribution scores. 
We then select the top-$k$ high-attribution spatio-temporal regions $\mathcal{R}=\{r_i\}_{i=1}^{k}$, where each region is described by its sensor channel, temporal interval, and average attribution score. 
These regions provide evidence candidates for observation nodes in the teacher trace.

\paragraph{Attribution Serialization.}
Each region $r_i\in\mathcal{R}$ is converted into a discrete evidence record before being passed to the teacher LLM. 
We map sensor channels to a dataset-specific closed vocabulary, quantize temporal intervals into coarse temporal phases, and convert attribution strengths into ordinal confidence levels based on rank-normalized saliency. 
The serialized records and their attribution scores are included in the teacher prompt.
This closed format reduces ambiguity in generated observation nodes and makes the resulting traces easier to parse and evaluate.

\paragraph{Structured Reasoning Generation.}
Conditioned on the serialized evidence records, the teacher LLM $\mathcal{T}$ generates a structured reasoning trace for each training sample. The teacher is constrained to follow a DAG schema that decomposes reasoning into four complementary stages: Observation, Inference, Synthesis, and Activity, with edges explicitly encoding evidence dependencies between them. Observation nodes ground the reasoning in attribution-supported sensor evidence (e.g., periodic arm acceleration), inference nodes combine related observations to derive localized intermediate conclusions (e.g., rhythmic arm swings), the synthesis node integrates multiple inferences into a coherent global explanation (e.g., coordinated upper- and lower-body movement), and the activity node represents the final prediction (e.g., walking upstairs). Each higher-level node includes explicit \texttt{based\_on} links to its supporting lower-level nodes, forming a provenance chain from raw sensor evidence to the final decision. By separating perception, localized reasoning, global reasoning, and decision making, the DAG enables stepwise verification and fine-grained error localization. The resulting traces are used as supervision targets for the student model in Stage~2.

\subsection{Stage 2: Cross-Modal Alignment}
The second stage trains a compact sensor-conditioned language model to generate the teacher-produced reasoning traces from raw sensor inputs. 
The time-series encoder and the student language model are kept frozen, while a projector and gated cross-attention adapters are trained for sensor-language alignment.

\paragraph{Sensor Encoding.}
Given a sensor window $\mathbf{X}$, we use a pre-trained time-series encoder $\mathcal{E}$ to extract channel-level embeddings. 
For $C$ sensor channels, the encoder produces embeddings that are concatenated into $\mathbf{z}\in\mathbb{R}^{D}$, where $D=C\cdot d_{\mathrm{ch}}$. 
A learned projector $g_\phi$ maps $\mathbf{z}$ into a fixed number of sensor memory tokens,
$
\mathbf{Z}=g_\phi(\mathbf{z})\in\mathbb{R}^{N\times d},
$
where $N$ is the number of memory tokens and $d$ matches the hidden dimension of the student model. 
These tokens provide a compact representation of the sensor window for cross-attention.

\paragraph{Gated Cross-Attention Adapters.}
To condition the frozen student model on sensor memory, we insert gated cross-attention adapters into each decoder layer \cite{alayrac2022flamingo, langer2025opentslm}. 
At layer $l$, decoder hidden states $\mathbf{H}^{(l-1)}$ produce queries $\mathbf{Q}^{(l)}$, while keys and values $\mathbf{K}^{(l)},\mathbf{V}^{(l)}$ are derived from the sensor memory tokens $\mathbf{Z}$. 
The layer output is updated as:
$
\mathbf{H}^{(l)}
=
\mathbf{H}^{(l-1)}
+
\alpha_l\,
\mathrm{CrossAttn}
\left(
\mathbf{Q}^{(l)}, \mathbf{K}^{(l)}, \mathbf{V}^{(l)}
\right)
$,
where $\alpha_l$ is a learnable gate initialized near zero. 
This initialization preserves the pretrained language model behavior at the start of training while allowing sensor-conditioned generation to emerge gradually. 
Only the projector $g_\phi$ and cross-attention adapters are updated; the time-series encoder and student language model remain frozen.

\paragraph{Training Objective.}
The student is trained to generate the teacher trace with a causal language modeling objective.
Each sequence consists of a fixed instruction prompt $p$ and a target reasoning trace $\mathbf{g}=(g_1,\ldots,g_M)$, with the activity label included in the trace.
The loss is applied only to the target trace tokens:
\begin{equation}
\mathcal{L}
=
-\sum_{m=1}^{M}
\log P(g_m \mid p, g_{<m}, \mathbf{Z}).
\label{eq:loss}
\end{equation}
Thus, the model learns to generate the activity prediction and the structured reasoning trace with a single autoregressive objective.

\subsection{Inference}
\label{sec:method:inference}

At inference, \textsc{TRACE-TS} no longer uses the expert classifier $\mathcal{C}$, teacher LLM $\mathcal{T}$, or attribution computation. 
Given a sensor window $\mathbf{X}$, the frozen encoder $\mathcal{E}$ and trained projector $g_\phi$ produce sensor memory tokens $\mathbf{Z}$, which condition the adapted student language model through the learned cross-attention adapters. 
The model then autoregressively generates a serialized reasoning graph $\mathcal{G}$ with the activity label $y$ in a single inference pass.

\begin{table*}[t]
\caption{F1-macro and accuracy scores (\%) for the Human Activity Recognition tasks across seven benchmark datasets. Bold indicates the best result, while underline denotes the second-best.}
\label{tab:cot_har_ours_comparison}
\centering
\scriptsize
\resizebox{\textwidth}{!}{%
\begin{tabular}{l|cc|cc|cc|cc|cc|cc|cc|cc}
\toprule
& \multicolumn{2}{c|}{UCI-HAR}
& \multicolumn{2}{c|}{USC-HAD}
& \multicolumn{2}{c|}{PAMAP2}
& \multicolumn{2}{c|}{CAPTURE-24}
& \multicolumn{2}{c|}{MHEALTH}
& \multicolumn{2}{c|}{SHOAIB}
& \multicolumn{2}{c|}{OPPORTUNITY}
& \multicolumn{2}{c}{Average} \\
Method & Acc & F1 & Acc & F1 & Acc & F1 & Acc & F1 & Acc & F1 & Acc & F1 & Acc & F1 & Acc & F1 \\
\midrule
\multicolumn{17}{l}{\textbf{HAR Baselines}} \\
\midrule
Attend \& Discriminate & \textbf{97.50} & \textbf{97.55} & 72.19 & \underline{73.24} & \underline{86.25} & \textbf{84.23} & 62.06 & 57.77 & 72.11 & 70.57 & 93.08 & 92.64 & \underline{77.48} & \underline{69.76} & 80.10 & 77.97 \\
MantisV2 & \underline{96.81} & \underline{96.97} & \textbf{73.09} & \textbf{77.22} & 70.33 & 63.98 & 60.34 & 58.97 & 83.93 & 67.29 & 99.14 & 99.14 & \textbf{84.33} & \textbf{77.77} & 81.14 & 77.33 \\
UniMTS & 90.97 & 91.05 & 60.23 & 64.57 & 68.28 & 59.24 & 57.46 & 59.20 & 80.42 & 55.25 & 97.38 & 97.36 & 31.07 & 23.25 & 69.40 & 64.27 \\
Chronos-2 & 89.69 & 90.08 & 61.70 & 62.73 & 78.93 & 71.82 & 62.84 & 59.94 & 76.64 & 73.11 & 96.67 & 96.66 & 52.70 & 41.30 & 74.17 & 70.81 \\
NST & 90.34 & 90.40 & 45.89 & 46.06 & 76.14 & 68.68 & 59.75 & 52.58 & 75.75 & 73.47 & 78.10 & 75.99 & 61.61 & 53.82 & 69.65 & 65.86 \\
PatchTST & 77.52 & 78.20 & 50.46 & 53.73 & 63.20 & 53.33 & 58.19 & 53.10 & 67.36 & 61.14 & 91.96 & 91.92 & 43.97 & 34.20 & 64.67 & 60.80 \\
\midrule
\multicolumn{17}{l}{\textbf{LLM-Based Methods}} \\
\midrule
SensorLLM             & 94.67 & 94.66 & 33.01 & 25.90 & 70.86 & 61.25 & \underline{63.95} & \textbf{61.09} & 98.10 & 98.22 & 97.38 & 97.37 &  9.11 &  4.49 & 66.73 & 63.28 \\
OpenTSLM (Fine-tuned) & 87.24 & 87.31 & 36.24 & 26.02 & 63.26 & 55.84 & 12.77 & 17.09 & 77.07 & 76.02 & 88.76 & 89.01 & 17.80 & 10.30 & 54.73 & 51.66 \\
TimeMQA (Qwen 2.5)    & 19.25 & 10.53 & 14.27 &  3.13 &  8.98 &  4.31 & 12.85 &  4.87 & 10.11 &  3.56 & 14.89 &  4.66 & 12.75 &  1.64 & 13.30 &  4.67 \\
LLaSA-7B              & 27.58 & 14.50 &  5.70 &  2.36 &  9.45 &  5.84 &  0.03 &  0.46 &  3.27 &  1.63 & 17.72 & 17.75 &  1.72 &  0.48 &  9.35 &  6.15 \\
GPT-OSS 120B (CoT)    & 35.45 & 27.91 & 17.04 &  9.82 & 16.38 & 13.03 & 54.23 & 31.34 & 31.42 & 23.06 & 27.34 & 23.55 &  6.98 &  3.80 & 26.98 & 18.93 \\
Llama 3.3 70B (CoT)   & 23.26 & 13.82 & 10.94 &  4.81 & 12.48 &  5.26 & 44.08 & 19.51 & 10.81 &  3.27 & 24.39 & 15.23 &  8.70 &  2.44 & 19.24 &  9.19 \\
Gemma 4 31B (CoT)     & 34.12 & 20.15 & 22.90 & 13.04 & 18.79 & 11.68 & 52.48 & 29.73 & 35.85 & 21.98 & 28.78 & 20.28 &  5.34 &  1.85 & 28.32 & 16.96 \\
Qwen 3.5 27B (CoT)    & 33.29 & 23.39 & 16.36 &  9.93 & 15.95 & 12.20 & 52.35 & 30.06 & 31.96 & 26.33 & 34.77 & 28.09 &  6.46 &  4.02 & 27.31 & 19.15 \\
\midrule
\multicolumn{17}{l}{\textbf{Ours (\textsc{TRACE-TS})}} \\
\midrule
Gemma 4 4B & 96.67 & 96.73 & 68.58 & 64.36 & \textbf{87.10} & \underline{83.08} & \textbf{63.96} & \underline{60.20} & 98.17 & 98.27 & \textbf{99.40} & \textbf{99.40} & 77.13 & 66.63 & \textbf{84.43} & \textbf{81.24} \\
Llama 3.2 3B & 96.20 & 96.25 & 70.83 & 65.43 & 85.50 & 79.10 & 63.41 & 58.81 & \textbf{99.19} & \textbf{99.23} & 99.25 & 99.24 & 72.42 & 62.23 & \underline{83.83} & 80.04 \\
Qwen 3.5 4B & 96.64 & 96.71 & \underline{72.35} & 67.88 & 84.55 & 78.64 & 63.56 & 59.92 & \underline{98.32} & \underline{98.38} & \underline{99.36} & \underline{99.36} & 71.70 & 61.08 & 83.78 & \underline{80.28} \\
\bottomrule
\end{tabular}%
}
\end{table*}

\section{Experiments}
\label{sec:evaluation}

\paragraph{Datasets.}
We evaluate \textsc{TRACE-TS} on seven wearable HAR benchmarks: UCI-HAR, USC-HAD, PAMAP2, CAPTURE-24, MHEALTH, SHOAIB, and OPPORTUNITY. 
These datasets cover controlled and free-living settings, vary from 3 to 79 sensor channels, and include activity labels with different levels of granularity. 
Details on dataset splits, preprocessing, activity classes, and sensor configurations are provided in Appendix \ref{sec:appendix_datasets}.

\paragraph{Baselines.}
We compare \textsc{TRACE-TS} with two groups of baselines. 
The first group includes time-series and sensor foundation models, including Chronos-2 \cite{ansari2025chronos2}, MantisV2 \cite{feofanov2026mantisv2}, PatchTST \cite{Yuqietal-2023-PatchTST}, NST \cite{liu2022non}, and UniMTS \cite{zhang2024unimts}, adapted to each dataset through fine-tuning or linear probing. 
The second group includes LLM-based time-series methods, including SensorLLM \cite{li2025sensorllm}, OpenTSLM \cite{langer2025opentslm}, TimeMQA \cite{kong-etal-2025-time}, LLaSA \cite{llasa} and direct CoT-prompting variants of Qwen~3.5 \cite{qwen3.5}, LLaMA~3.2/3.3 \cite{grattafiori2024llama}, Gemma~4 \cite{gemma4}, and GPT-OSS \cite{gptoss}.
This comparison covers both specialist HAR models and language-based reasoning approaches.

\paragraph{\textsc{TRACE-TS} Configuration.}
Unless otherwise specified, we use Gemma~4 4B \cite{gemma4} as the student language model, Qwen~3.5-122B-A10B as the teacher LLM, MantisV2 as the frozen sensor encoder, and Attend-and-Discriminate as the expert HAR classifier. Qwen 3.5 35B-A3B serves as the primary judge for the SNM metric (Sec.~\ref{sec:evaluation_metrics}).
We use $N{=}8$ sensor memory tokens and gated cross-attention adapters with rank $r{=}128$ at each decoder layer. 
Ablations over student models, sensor token count, and adapter rank appear in Sec.~\ref{sec:ablation}, with implementation details in Appendix~\ref{sec:appendix_impl_trace}.

\subsection{Evaluation Metrics}
\label{sec:evaluation_metrics}

We evaluate \textsc{TRACE-TS} along two axes. 
For activity recognition, we report accuracy and macro-F1 on the test set. 
For reasoning quality, we compare standard NLG metrics with our proposed structure-aware metric, \emph{Semantic Node Match} (SNM).

SNM evaluates a generated reasoning trace against a reference DAG trace at three levels: observation, inference, and synthesis. 
For observation and inference nodes, we formulate node matching as an assignment problem between reference nodes $\mathcal{O}^*=\{o_1^*,\ldots,o_m^*\}$ and predicted nodes $\hat{\mathcal{O}}=\{\hat{o}_1,\ldots,\hat{o}_k\}$. 
The optimal matching $\pi^*=\operatorname*{arg\,max}_{\pi\in\Pi}\sum_{(i,j)\in\pi}w(o_i^*,\hat{o}_j)$ maximizes the total node consistency score and is solved with the Hungarian algorithm~\citep{kuhn1955hungarian}. 
The pairwise score $w(o_i^*,\hat{o}_j)$ combines structural matching and semantic matching:
\begin{equation}
w(o_i^*, \hat{o}_j) =
\begin{cases}
\begin{aligned}
&\mathcal{J}\bigl(\phi(o_i^*), \phi(\hat{o}_j)\bigr),\\[-1mm]
&\quad \text{if } \sigma(o_i^*) = \sigma(\hat{o}_j),
\end{aligned}\\
0, \quad \text{otherwise}.
\end{cases}
\label{eq:snm_gate}
\end{equation}
Here, $\sigma(\cdot)$ extracts the referenced sensor channel, $\phi(\cdot)$ extracts the node description, and $\mathcal{J}(\cdot,\cdot)$ is an LLM judge for semantic equivalence. 
We report \textbf{SNM-OF1} for observation-node matching, \textbf{SNM-IF1} for inference-node matching, and \textbf{SNM-SA} for sample-level agreement between predicted and reference synthesis nodes.

\begin{table}[t]
\small
\caption{NLG metrics for \textsc{TRACE-TS} (Gemma~4 4B) conditioned on classification correctness. \cmark/\xmark\ denote correct/incorrect predictions, with $\Delta{=}$\cmark${-}$\xmark.}
\label{tab:cot_conditioned_generation}
\centering
\setlength{\tabcolsep}{3pt}
\begin{tabular}{l|rrr|rrr}
\toprule
& \multicolumn{3}{c|}{BERTScore (\%)} & \multicolumn{3}{c}{METEOR (\%)} \\
Dataset & \cmark & \xmark & $\Delta$ & \cmark & \xmark & $\Delta$ \\
\midrule
UCI-HAR    & 95.89 & 95.60 & $+$0.29 & 61.80 & 59.38 & $+$2.42 \\
USC-HAD    & 95.88 & 95.78 & $+$0.10 & 62.83 & 58.25 & $+$4.58 \\
PAMAP2     & 95.31 & 95.09 & $+$0.22 & 57.77 & 55.82 & $+$1.95 \\
CAPTURE-24 & 95.97 & 95.78 & $+$0.19 & 62.97 & 58.56 & $+$4.41 \\
MHEALTH    & 95.30 & 95.55 & $-$0.25 & 58.02 & 56.60 & $+$1.42 \\
SHOAIB     & 95.64 & 95.08 & $+$0.56 & 59.76 & 56.04 & $+$3.72 \\
OPPORTUNITY       & 95.28 & 95.30 & $-$0.02 & 62.35 & 61.47 & $+$0.88 \\
\midrule
Average    & 95.61 & 95.45 & $+$0.16 & 60.79 & 58.02 & $+$2.77 \\
\bottomrule
\end{tabular}
\end{table}

\subsection{Main Results}
\label{sec:main_results}

\paragraph{Overall Recognition Performance.}
Table~\ref{tab:cot_har_ours_comparison} reports activity recognition results for \textsc{TRACE-TS} and representative baselines. 
Among the variants, \textsc{TRACE-TS} with Gemma~4 4B achieves the strongest average performance, improving over the best specialist HAR baseline by 3.29 accuracy points and 3.27 F1 points. 
It outperforms the strongest LLM-based baseline by 17.70 accuracy and 17.96 macro-F1 points. Additional baselines appear in Table \ref{tab:full_results} in Appendix~\ref{app:appendix_full_results}.

\paragraph{Performance across Datasets.}
Across most datasets, \textsc{TRACE-TS} ranks among the top-performing methods while also generating structured reasoning. 
USC-HAD and OPPORTUNITY are the main exceptions, where specialist HAR models remain competitive or stronger: USC-HAD due to high inter-subject variability under its leave-subjects-out protocol, and OPPORTUNITY due to fine-grained activity labels like \textit{Close Drawer 1} and \textit{Close Drawer 2}.
These results suggest that \textsc{TRACE-TS} preserves strong HAR performance while extending beyond label prediction.

\paragraph{Comparison with LLM-Based Methods.}
We also observe a clear gap between \textsc{TRACE-TS} and existing LLM-based time-series methods such as OpenTSLM. 
Although these methods use language models for time-series reasoning, their weaker performance suggests that effective sensor-language alignment depends not only on the language model, but also on the quality of the sensor encoder and how sensor evidence is injected into generation.
We also compare \textsc{TRACE-TS} with these baselines on the generated reasoning, using SNM (Appendix~\ref{app:snm_all_methods}) and conventional NLG metrics (Appendix~\ref{sec:appendix_generation}) and it substantially outperforms baselines under both.

\begin{table}[t]
\caption{Diagnostic SNM metrics for \textsc{TRACE-TS} (Gemma~4 4B), averaged across four LLM judges. \cmark/\xmark\ = correct/incorrect predictions; $\Delta{=}$\cmark${-}$\xmark. Values shown as mean$_{\pm\text{std}}$ (\%).}
\label{tab:snm_results}
\centering
\resizebox{\columnwidth}{!}{%
\begin{tabular}{l|ccc|ccc|ccc}
\toprule
& \multicolumn{3}{c|}{SNM-OF1 (\%)} & \multicolumn{3}{c|}{SNM-IF1 (\%)} & \multicolumn{3}{c}{SNM-SA (\%)} \\
Dataset & \cmark & \xmark & $\Delta$ & \cmark & \xmark & $\Delta$ & \cmark & \xmark & $\Delta$ \\
\midrule
UCI-HAR     & $40.4_{\pm9.6}$   & $38.6_{\pm9.4}$   & $+1.8_{\pm2.2}$   & $71.3_{\pm18.2}$ & $60.7_{\pm22.7}$ & $+10.6_{\pm6.9}$   & $92.9_{\pm9.3}$   & $14.9_{\pm10.6}$ & $+78.0_{\pm4.7}$ \\
USC-HAD     & $38.2_{\pm7.9}$   & $29.6_{\pm7.8}$   & $+8.6_{\pm1.4}$   & $68.4_{\pm19.1}$ & $44.5_{\pm17.1}$ & $+23.9_{\pm6.6}$   & $91.0_{\pm9.7}$   & $12.8_{\pm6.8}$  & $+78.3_{\pm9.0}$ \\
PAMAP2      & $15.5_{\pm4.8}$   & $14.2_{\pm4.5}$   & $+1.4_{\pm0.5}$   & $55.8_{\pm19.6}$ & $26.6_{\pm13.1}$ & $+29.2_{\pm7.0}$   & $74.5_{\pm7.6}$   & $5.3_{\pm0.9}$   & $+69.2_{\pm7.0}$ \\
CAPTURE-24  & $42.5_{\pm10.4}$  & $38.0_{\pm10.0}$  & $+4.5_{\pm2.0}$   & $71.4_{\pm16.5}$ & $20.0_{\pm11.8}$ & $+51.4_{\pm7.6}$   & $89.9_{\pm11.7}$  & $0.4_{\pm0.4}$   & $+89.6_{\pm11.4}$ \\
MHEALTH     & $33.2_{\pm9.0}$   & $31.3_{\pm7.6}$   & $+1.9_{\pm2.0}$   & $63.3_{\pm17.1}$ & $62.7_{\pm24.5}$ & $+0.6_{\pm11.8}$   & $90.0_{\pm8.6}$   & $53.5_{\pm24.8}$ & $+36.5_{\pm18.2}$ \\
SHOAIB      & $26.5_{\pm8.8}$   & $13.2_{\pm4.8}$   & $+13.3_{\pm5.0}$  & $66.2_{\pm20.7}$ & $35.4_{\pm23.7}$ & $+30.9_{\pm4.6}$   & $94.9_{\pm5.2}$   & $0.0_{\pm0.0}$   & $+94.9_{\pm5.2}$ \\
OPPORTUNITY & $19.4_{\pm5.4}$   & $19.0_{\pm5.8}$   & $+0.4_{\pm0.4}$   & $53.2_{\pm22.6}$ & $41.9_{\pm21.4}$ & $+11.3_{\pm2.1}$   & $87.5_{\pm8.5}$   & $16.4_{\pm7.0}$  & $+71.2_{\pm5.2}$ \\
\midrule
Average     & $30.8_{\pm8.0}$   & $26.3_{\pm7.1}$   & $+4.6_{\pm2.0}$   & $64.2_{\pm19.1}$ & $41.7_{\pm19.2}$ & $+22.5_{\pm6.7}$   & $88.7_{\pm8.7}$   & $14.8_{\pm7.2}$  & $+73.9_{\pm8.7}$ \\
\bottomrule
\end{tabular}%
}
\end{table}

\begin{figure*}
    \centering
    \includegraphics[width=\linewidth]{figures/qualitative.pdf}
    \Description{Video frames of a person opening a drawer alongside sensor traces and the reasoning outputs of TRACE-TS, OpenTSLM, TimeMQA, and SensorLLM, with correct reasoning highlighted in green and mistakes in red.}
    \caption{Qualitative reasoning examples and comparisons. We use green to highlight
correct parts, red for mistakes. Video is only used for qualitative verification.}
    \label{fig:qualitative}
\end{figure*}

\subsection{Reasoning Evaluation}
\label{sec:reasoning_eval}

\paragraph{Limitations of Standard NLG Metrics.}
Reasoning quality is harder to evaluate than activity recognition because there is no universally accepted metric for sensor-grounded explanations. 
We first examine standard NLG metrics in Table~\ref{tab:cot_conditioned_generation}. 
BERTScore and METEOR show only small gaps between correctly and incorrectly classified samples. 
Across datasets, the largest BERTScore gap is only 0.56 points, and incorrectly classified samples sometimes obtain higher BERTScores than correct ones.
METEOR shows slightly larger gaps, but the maximum gap is still only 4.58 points. 
These results suggest that surface-level text similarity is insufficient for diagnosing whether a reasoning trace follows the expected sensor-evidence structure.

\paragraph{Structure-Aware Evaluation with SNM}
Table~\ref{tab:snm_results} reports SNM-OF1, SNM-IF1, and SNM-SA averaged across four LLM judges (Qwen3.5-35B, Gemma-4-31B, GPT-OSS-120B, LLaMA-3.3-70B).
Compared to NLG metrics, SNM provides substantially clearer separation between traces associated with correct and incorrect predictions: averaged across the four judges, the gaps reach 4.6, 22.5, and 73.9 points for SNM-OF1, SNM-IF1, and SNM-SA respectively, with per-judge breakdowns in Appendix~\ref{app:snm_judge}.
The cross-judge spread (e.g., $\pm$19 on SNM-IF1) reflects calibration differences across judges, while rankings remain stable across all four, supporting the robustness of the metric.
The larger gaps at inference and synthesis levels suggest that reasoning failures arise primarily from composing observations into higher-level conclusions rather than from observation generation alone.

\paragraph{Dataset-level Variation.}
The SNM results also reveal differences across datasets. 
Datasets with fewer channels, such as USC-HAD and CAPTURE-24, tend to show clearer separation between correct and incorrect traces, likely because there are fewer plausible evidence paths. 
In contrast, PAMAP2 and OPPORTUNITY contain more sensors and more fine-grained activities, allowing multiple observation paths to support similar high-level reasoning. 
As a result, observation-level matching can be stricter and less stable, while inference- and synthesis-level metrics provide a more informative view of reasoning consistency.
\begin{table}[t]
\scriptsize
\caption{Dataset-wise expert ratings from five annotators (mean $\pm$ SD, 1--5 scale) for soundness, grounding, and overall quality.}
\label{tab:expert-eval}
\centering
\resizebox{\columnwidth}{!}{%
\begin{tabular}{lccc}
\toprule
Dataset & Soundness & Grounding & Overall \\
\midrule
UCI-HAR     & $4.72 \pm 0.53$ & $4.60 \pm 0.72$ & $4.63 \pm 0.67$ \\
USC-HAD     & $4.20 \pm 1.09$ & $4.03 \pm 1.10$ & $4.13 \pm 1.07$ \\
PAMAP2      & $4.22 \pm 1.17$ & $3.78 \pm 1.12$ & $3.95 \pm 1.19$ \\
CAPTURE-24  & $4.24 \pm 0.91$ & $4.38 \pm 0.78$ & $4.34 \pm 0.77$ \\
MHEALTH     & $4.45 \pm 0.75$ & $4.40 \pm 0.79$ & $4.43 \pm 0.70$ \\
SHOAIB      & $4.66 \pm 0.84$ & $4.51 \pm 0.82$ & $4.69 \pm 0.76$ \\
OPPORTUNITY & $4.26 \pm 0.95$ & $3.94 \pm 1.06$ & $4.11 \pm 0.93$ \\
\midrule
All         & $4.35 \pm 0.95$ & $4.15 \pm 1.00$ & $4.26 \pm 0.95$ \\
\bottomrule
\end{tabular}%
}
\end{table}

\paragraph{Human Evaluation.} Since automatic metrics cannot fully capture the quality of generated reasoning, we additionally conduct a human evaluation, with aggregated results reported in Table~\ref{tab:expert-eval}. We randomly sample one instance from each activity class across all seven datasets, and five human experts independently rate each reasoning trace on a five-point Likert scale for \textit{(i) Soundness}, assessing whether the reasoning is logically coherent; \textit{(ii) Grounding}, assessing whether the cited evidence is faithfully supported by the input sensor signal; and \textit{(iii) Overall Quality}, providing a holistic assessment of the explanation. Table~\ref{tab:expert-eval} shows that \textsc{TRACE-TS} consistently produces high-quality reasoning across datasets, achieving average scores of \textbf{4.35} for soundness, \textbf{4.15} for grounding, and \textbf{4.26} overall. While soundness is consistently rated slightly higher than grounding, this gap is expected: a prediction may be logically justified yet still reference unsupported evidence. For example, a trace may correctly predict \textit{walking} from rhythmic leg and arm motion, but if the input contains no rhythmic arm motion, the fabricated observation is penalized under grounding despite the correct prediction. Conversely, reasoning that is logically coherent and faithfully grounded receives partial credit even when the predicted activity is incorrect, allowing the evaluation to assess reasoning quality independently of classification accuracy. Representative examples are provided in Appendix~\ref{app:error_analysis}.

\paragraph{Qualitative Results.}
We also present a qualitative comparison on OPPORTUNITY++~\citep{ciliberto2021opportunity}, which provides synchronized video alongside time-series sensor data. This paired modality allows us to qualitatively cross-check whether reasoning aligns with the visible activity. As shown in Figure~\ref{fig:qualitative}, baseline models produce plausible but unsupported rationales: OpenTSLM predicts \textit{walking}, TimeMQA predicts \textit{Clean Table}, and SensorLLM mainly describes low-level signal trends without identifying the activity. In contrast, \textsc{TRACE-TS} correctly predicts \textit{Open Drawer} and identifies motion cues such as \textbf{twisting of the trunk} and \textbf{shifted posture}, both visible in the video. It further infers that the person opens a drawer below waist level, which is consistent with the final video frame.

\subsection{Ablation and Analysis}
\label{sec:ablation}

To understand the role of each design choice in \textsc{TRACE-TS}, we perform a series of controlled ablation studies in which each component is individually varied while all others remain fixed, measuring its effect on both predictive performance and reasoning quality. 

\paragraph{Effect of Attribution Source.}
We evaluate the importance of attribution quality by training five variants across all seven datasets using different attribution settings: SHAP only, IG only, randomly shuffled attributions, no attribution (No-XAI), and conditioning on only the 10\% least-salient channels. Table~\ref{tab:attr_ablation} reports the aggregate results across all seven datasets, while detailed per-dataset results are provided in Appendix~\ref{app:attr_ablation}. Replacing the fused attribution with SHAP or IG alone reduces classification accuracy by \textbf{1.81}\% and \textbf{2.83}\%, respectively. Random attributions, no attribution, and least-salient channels incur drops of \textbf{2.23}\%, \textbf{3.73}\%, and \textbf{4.52}\%, confirming that high-quality attribution signals provide stronger supervision for learning grounded reasoning. To further understand how attribution quality affects the learned reasoning, we additionally evaluate every variant using SNM. Among the faithful attribution sources (SHAP, IG, and the fused attribution), SNM scores remain comparable, with differences falling within normal run-to-run variation. In contrast, degraded attribution regimes substantially weaken reasoning quality despite only modest accuracy drops, with random attribution reducing observation grounding by \textbf{12.78} points (from \textbf{21.49} to \textbf{8.71} SNM-OF1) and removing attribution entirely eliminating observation grounding (\textbf{SNM-OF1 = 0.00}), while reducing synthesis agreement by \textbf{29.57} points (from \textbf{76.14} to \textbf{46.57}). To verify that this degradation is not an artifact of scoring against attribution-conditioned references, we conduct a blind pairwise human evaluation comparing traces generated with the fused attribution against those of the no-attribution variant: five annotators, blind to the generating condition and with randomized presentation order, prefer the attribution-grounded trace in \textbf{82.4}\% of all judgments and \textbf{91.2}\% of decisive judgments (Table~\ref{tab:human_ab}, Appendix~\ref{app:attr_ablation}). Together, these findings demonstrate the critical role of high-quality attribution in generating reasoning traces grounded in the sensor evidence.
\begin{table}[t]
\caption{Ablation results for different attribution sources, averaged across seven datasets. Parenthesized values denote performance differences w.r.t. original settings.}
\label{tab:attr_ablation}
\centering
\small
\setlength{\tabcolsep}{5pt}
\resizebox{\columnwidth}{!}{%
\begin{tabular}{l|cc|ccc}
\toprule
& \multicolumn{2}{c|}{\textbf{Classification}} & \multicolumn{3}{c}{\textbf{Reasoning quality}} \\
Method & Average Accuracy & Average F1 & SNM-OF1 & SNM-IF1 & SNM-SA \\
\midrule
SHAP + IG (Original) & 84.43 & 81.24 & 21.49 & 39.16 & 76.14 \\
SHAP-only          & 82.62 ($-$1.81) & 80.85 ($-$0.39) & 19.44 ($-$2.05) & 38.52 ($-$0.64) & 76.19 ($+$0.05) \\
IG-only            & 81.60 ($-$2.83) & 79.52 ($-$1.72) & 21.86 ($+$0.37) & 37.98 ($-$1.18) & 75.25 ($-$0.89) \\
Random attribution & 82.20 ($-$2.23) & 80.33 ($-$0.91) &  8.71 ($-$12.78) & 34.13 ($-$5.03) & 73.63 ($-$2.51) \\
No attribution             & 80.70 ($-$3.73) & 78.27 ($-$2.97) &  0.00 ($-$21.49) & 18.62 ($-$20.54) & 46.57 ($-$29.57) \\
Bottom-10\% attr.  & 79.91 ($-$4.52) & 77.97 ($-$3.27) &  5.00 ($-$16.49) & 31.65 ($-$7.51) & 69.39 ($-$6.75) \\
\bottomrule
\end{tabular}
}
\end{table}

\paragraph{Effect of Reasoning Format.}
To isolate the contribution of the structured DAG output format in \textsc{TRACE-TS}, we train three Gemma~4 4B ablation variants derived from the same attribution-conditioned teacher traces: \textit{Label-only}, which discards the reasoning trace and uses only the activity label; \textit{Free form output}, which removes node identifiers, node types, and \texttt{based\_on} edges and flattens the trace into paragraph-style reasoning; and \textit{CoT}, which preserves the observation--inference--synthesis--answer order but removes graph-level provenance. As shown in Table~\ref{tab:output_ablation}, label-only supervision, free-form reasoning, and CoT lead to average drops of \textbf{5.46}\%, \textbf{20.57}\%, and \textbf{22.93}\% in accuracy, and \textbf{5.04}\%, \textbf{19.58}\%, and \textbf{22.38}\% in macro-F1, respectively. These results indicate that unstructured reasoning is detrimental as a supervision signal, whereas the DAG structure and provenance links enable reasoning to improve, rather than hinder, classification. Per-dataset results are provided in Appendix~\ref{app:output_ablation}.

\begin{table}[t]
\caption{Ablation results for different reasoning formats, averaged across seven datasets. Parenthesized values denote performance differences w.r.t. original setting.}
\label{tab:output_ablation}
\centering
\small
\begin{tabular}{lcc}
\toprule
Method & Average Accuracy & Average F1 \\
\midrule
DAG (Original)   & 84.43 & 81.24 \\
Label-only       & 78.97 ($-$5.46)  & 76.20 ($-$5.04) \\
Free form output & 63.86 ($-$20.57) & 61.66 ($-$19.58) \\
CoT reasoning    & 61.50 ($-$22.93) & 58.86 ($-$22.38) \\
\bottomrule
\end{tabular}
\end{table}

\paragraph{Effect of Sensor Encoder and Expert Classifier.}
We assess the role of the upstream components by replacing the default MantisV2 sensor encoder with MOMENT \cite{goswami2024moment} and the Attend-and-Discriminate expert classifier with DeepConvLSTM \cite{ordonez2016deep}, keeping all other components fixed. As shown in Table~\ref{tab:encoder_clf_ablation} and Appendix~\ref{app:encoder_clf_ablation}, replacing the sensor encoder with MOMENT reduces average accuracy and macro-F1 by \textbf{19.48}\% and \textbf{19.81}\%, respectively, while replacing the expert classifier with DeepConvLSTM reduces them by \textbf{10.63}\% and \textbf{10.89}\%. These results highlight the importance of high-quality upstream components, particularly the sensor encoder, whose representations are critical for cross-attention alignment and reasoning.  

\begin{table}[t]
\caption{Ablation results for different sensor encoders \& expert classifiers, averaged across seven datasets. Parenthesized values denote differences w.r.t. the original setting.}
\label{tab:encoder_clf_ablation}
\centering
\scriptsize
\setlength{\tabcolsep}{3.5pt}
\begin{tabular}{lcc}
\toprule
Method & Average Accuracy & Average F1 \\
\midrule
Mantis + Attend (Original)   & 84.43 & 81.24 \\
Moment + Attend                  & 64.95 ($-$19.48) & 61.43 ($-$19.81) \\
Mantis + DeepConvLSTM        & 73.80 ($-$10.63) & 70.35 ($-$10.89) \\
\bottomrule
\end{tabular}
\end{table}

\paragraph{Reasoning Trace Perturbation.}
To further evaluate the robustness of SNM, we conduct two controlled perturbation studies on the reference reasoning traces and compare its behavior against conventional NLG metrics. We design the perturbations to independently test robustness to surface-level rewording and sensitivity to corrupted evidence grounding. The first perturbation paraphrases each observation and inference while preserving the underlying sensor claims and provenance edges. The second preserves the surface text but replaces the cited channels and \texttt{based\_on} identifiers with plausible but incorrect ones, thereby misgrounding the reasoning trace in incorrect evidence. As shown in Table~\ref{tab:perturbation_ablation} and Appendix~\ref{app:perturbation_ablation}, \textit{Paraphrase} reduces METEOR by \textbf{4.63}\% while leaving SNM scores largely unchanged, demonstrating robustness to surface level paraphrasing. In contrast, \textit{Misgrounded} has only a minor effect on text metrics (BS $-$0.49, MET $-$6.08) but reduces SNM-IF1 by \textbf{24.35}\% and SNM-SA by \textbf{72.01}\%, demonstrating SNM's sensitivity to corrupted evidence grounding. These findings demonstrate that SNM measures whether reasoning remains grounded in the underlying sensor evidence rather than merely capturing textual similarity. 
\begin{table}[t]
\caption{Perturbation ablation on the reference reasoning traces, averaged across seven datasets. All values in (\%); parentheses denote change w.r.t.\ the unperturbed reference. $^{\dagger}$Macro-F1 is unchanged: perturbation affects only the reference traces, not model predictions.}
\label{tab:perturbation_ablation}
\centering
\small
\resizebox{\columnwidth}{!}{%
\begin{tabular}{lcccccc}
\toprule
Reference & Macro-F1$^{\dagger}$ & BERTScore & METEOR & SNM-OF1 & SNM-IF1 & SNM-SA \\
\midrule
Original (TRACE-TS) & 81.24 & 95.64 & 61.01 & 21.49 & 39.16 & 76.14 \\
Paraphrase          & 81.24 & 95.08 ($-$0.56) & 56.38 ($-$4.63) & 22.50 ($+$1.01) & 37.88 ($-$1.28)  & 75.22 ($-$0.92)  \\
Misgrounded         & 81.24 & 95.15 ($-$0.49) & 54.93 ($-$6.08) & 22.67 ($+$1.18) & 14.81 ($-$24.35) &  4.13 ($-$72.01) \\
\bottomrule
\end{tabular}%
}
\end{table}

\begin{figure}[t]
  \centering
  \includegraphics[width=\linewidth]{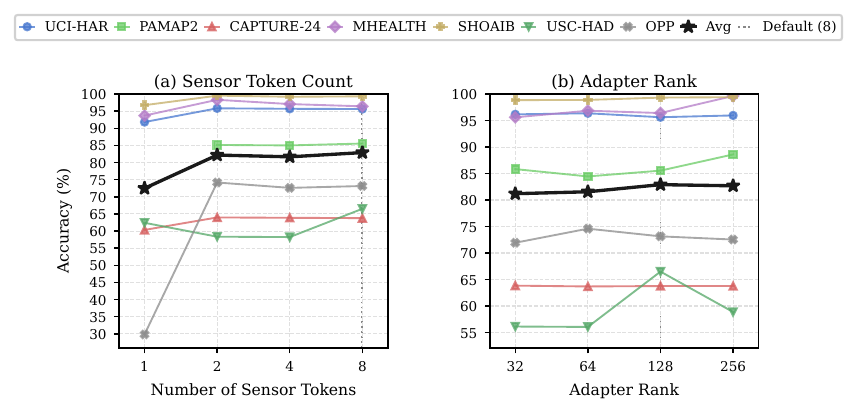}
  \Description{Line charts of accuracy on the seven datasets as the number of sensor tokens varies from 1 to 8 and the adapter rank varies from 32 to 256.}
  \caption{Ablation analysis of (a) sensor token count and (b) adapter rank on model performance}
  \label{fig:ablation}
\end{figure}
\begin{figure}[t]
  \centering
  \includegraphics[width=\linewidth]{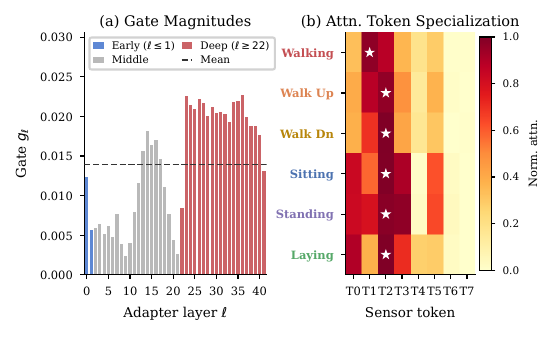}
  \Description{Bar chart of adapter gate magnitudes across decoder layers, which grow in deeper layers, and a heatmap showing different sensor memory tokens attending to different activity classes.}
  \caption{(a) Layer-wise adapter gating and (b) sensor-token cross-attention patterns.}
  \label{fig:attribution}
\end{figure}

\paragraph{Effect of Sensor Memory Size and Adapter Rank.}
We study the effect of sensor token count and adapter rank in Fig.~\ref{fig:ablation}. 
Increasing the number of sensor memory tokens generally improves performance, especially on higher-channel datasets such as PAMAP2 and OPPORTUNITY. 
Performance saturates with a small number of tokens, suggesting that a compact sensor memory is sufficient for effective sensor-conditioned generation. 
Varying the adapter rank from 32 to 256 leads to relatively small fluctuations, indicating that \textsc{TRACE-TS} does not require large adaptation capacity to align sensor representations with the language model.

\paragraph{Effect of Backbone Architecture.}

We assess the role of teacher quality by replacing the default Qwen~3.5-122B teacher with LLaMA~3.3-70B-Instruct, Gemma~4-31B, and GPT-OSS-120B, while keeping all other settings fixed. As shown in Table~\ref{tab:teacher_comparison}, LLaMA~3.3-70B, Gemma~4-31B, and GPT-OSS-120B lead to average drops of \textbf{4.65}\%, \textbf{4.65}\%, and \textbf{2.35}\% in accuracy, and \textbf{4.42}\%, \textbf{3.84}\%, and \textbf{1.62}\% in macro-F1, respectively. The consistent improvement with stronger teachers underscores the importance of high-quality reasoning traces for effective supervision. In contrast, changing the student backbone from Gemma~4 4B to equivalently sized Qwen and LLaMA variants results in only minor variations of \textbf{0.65\%}/\textbf{1.20\%} in average Accuracy/F1 (Table~\ref{tab:cot_har_ours_comparison}), indicating that the framework remains largely student backbone-agnostic while remaining sensitive to teacher capacity.

\begin{table}[t]
\caption{Effect of the teacher model on the performance of \textsc{TRACE-TS}, averaged across seven datasets.}
\label{tab:teacher_comparison}
\centering
\begin{tabular}{lcc}
\toprule
Teacher & Avg Acc & Avg F1 \\
\midrule
LLaMA 3.3 70B   & 79.78 & 76.82 \\
Gemma 4 31B     & 79.78 & 77.40 \\
GPT-OSS 120B    & 82.08 & 79.62 \\
Qwen 3.5 122B   & 84.43 & 81.24 \\
\bottomrule
\end{tabular}
\end{table}
\paragraph{Gating and Token Specialization.}
We further analyze the learned cross-attention behavior in Fig.~\ref{fig:attribution}. 
Adapter gates are smaller in early and middle layers and become stronger in deeper layers, suggesting that sensor information is used more heavily during later stages of generation \cite{koval-etal-2025-multimodal, zhang2025unravelling}. 
The attention maps also reveal token specialization, with different sensor memory tokens attending to complementary activity-discriminative patterns. This provides qualitative evidence that the model uses sensor memory in a structured way rather than injecting uniform information across layers.

\section{Conclusion}

We introduced \textsc{TRACE-TS}, a sensor-language reasoning framework that generates activity predictions together with structured reasoning traces for wearable sensor data. 
\textsc{TRACE-TS} uses expert-classifier attribution to construct evidence-grounded supervision, organizes reasoning as DAG traces with explicit provenance links, and distills this structure into a compact sensor-conditioned language model through gated cross-attention adapters.
Across seven HAR benchmarks, \textsc{TRACE-TS} achieves strong activity recognition performance while producing traceable reasoning outputs. 
We further show that standard NLG metrics provide limited diagnostic value for sensor-grounded reasoning and traceability, and introduce Semantic Node Match (SNM) to compare traces at the observation, inference, and synthesis levels.
Overall, our results suggest that attribution-grounded and structured reasoning is a promising direction for moving wearable sensor models beyond label prediction toward more interpretable sensor-language understanding.

\section{Limitations}

\textsc{TRACE-TS} relies on teacher-generated reasoning traces for supervision. Although these traces are constructed from attribution-supported sensor regions, their quality is still bounded by the teacher LLM and the attribution pipeline. Errors or omissions in the teacher traces may therefore be inherited by the student model. Similarly, SNM evaluates generated traces against teacher-generated reference traces, so it should be viewed as a structure-aware diagnostic metric rather than an absolute verification of reasoning faithfulness. Multi-sensor inputs may also admit multiple plausible reasoning paths, causing structurally different but semantically reasonable traces to receive lower SNM scores. Finally, while \textsc{TRACE-TS} is designed as a modular framework for grounded reasoning over multivariate time series, our evaluation is limited to wearable sensor datasets for human activity recognition. Although its modular architecture is intended to support extension to other multivariate time-series domains, such as healthcare monitoring, industrial sensing, and multimodal physiological analysis, rigorous empirical validation in these settings is beyond the scope of this work and remains an important direction for future research.

\begin{acks}
The authors gratefully acknowledge the research infrastructure and computational facilities provided by Thapar Institute of Engineering and Technology, Patiala, India, which enabled this work.
\end{acks}

\bibliographystyle{ACM-Reference-Format}
\bibliography{custom}

\clearpage
\appendix
\section{Datasets}
\label{sec:appendix_datasets}
We used seven datasets in our study. All datasets are publicly available, containing no personally identifiable information, thus posing minimal ethical or privacy concerns. Per-dataset license names, sources, and terms of use, together with the licenses of the pretrained models, are provided in Appendix~\ref{sec:appendix_licensing}.

\paragraph{Opportunity \citep{roggen2010collecting}.}
The Opportunity dataset consists of body-worn sensor data collected from four subjects performing a series of morning kitchen activities. The recordings include triaxial accelerometer, gyroscope, and magnetometer streams from five on-body IMUs (back, right upper arm, right lower arm, left upper arm, and left lower arm, with the back IMU additionally contributing Quat1 and Quat2), together with 16 channels each from the L-Shoe and R-Shoe InertiaCube3 sensors (Euler angles, navigation-frame and body-frame accelerations, body-frame and navigation-frame angular velocities, and compass), yielding 79 input channels selected from the original 250-column raw recordings. The data is sampled at 30~Hz and includes 18 activity class labels (17 mid-level gestures plus a Null class). For evaluation, we follow the standard split used in \citet{abedin2021attend}: S2-ADL4, S2-ADL5, S3-ADL4, and S3-ADL5 form the test set, with the remaining recordings used for training. A window size $w = 24$ with a stride of 12 ($\sim$0.80~s, 50\% overlap) is used.

\paragraph{Opportunity++ \citep{ciliberto2021opportunity}.}
Opportunity++ is a multimodal re-release of the original Opportunity dataset that additionally provides synchronized video footage, object and ambient sensor streams, and refined skeleton-tracking annotations alongside the body-worn inertial recordings. The wearable sensor data, activity labels, sampling rate, channel selection, and windowing are identical to those of Opportunity, so we apply the exact same preprocessing pipeline described above. We do not train on Opportunity++; we use it for inference only in our qualitative analysis (Section~\ref{sec:reasoning_eval}), where the released synchronized video lets us cross-reference and visually verify the activity segments underlying model-generated reasoning traces.

\begin{table*}[ht]
\centering
\footnotesize
\caption{Class distribution statistics across all seven HAR benchmark datasets, aggregated over the train, validation, and test splits}
\label{tab:dataset_proportions}
\renewcommand{\arraystretch}{1.25}
\setlength{\tabcolsep}{4pt}
\begin{tabularx}{\textwidth}{@{} l c X p{4.2cm} @{}}
\toprule
\textbf{Dataset} & \textbf{\# Classes} & \textbf{Classes} & \textbf{Proportions (\%)} \\
\midrule
Capture24 & 6 & Sleep, Sit-stand, Walking, Bicycling, Mixed, Vehicle & 30.45, 30.23, 12.25, 1.84, 18.63, 6.61 \\
\midrule
Opportunity & 18 & Null, Close Dishwasher, Close Drawer 3, Close Drawer 2, Close Door 1, Close Door 2, Close Drawer 1, Close Fridge, Toggle Switch, Open Dishwasher, Open Drawer 3, Open Drawer 2, Open Door 1, Open Door 2, Open Drawer 1, Open Fridge, Drink from Cup, Clean Table & 76.66, 1.02, 0.86, 0.58, 1.21, 1.26, 0.62, 1.79, 1.04, 1.10, 0.89, 0.68, 1.33, 1.37, 0.75, 1.98, 5.28, 1.59 \\
\midrule
Opportunity (no-null) & 17 & Close Dishwasher, Close Drawer 3, Close Drawer 2, Close Door 1, Close Door 2, Close Drawer 1, Close Fridge, Toggle Switch, Open Dishwasher, Open Drawer 3, Open Drawer 2, Open Door 1, Open Door 2, Open Drawer 1, Open Fridge, Drink from Cup, Clean Table & 4.43, 3.71, 2.43, 5.19, 5.41, 2.62, 7.64, 4.44, 4.70, 3.75, 2.95, 5.73, 5.85, 3.21, 8.50, 22.66, 6.77 \\
\midrule
PAMAP2 & 12 & Rope Jumping, Lying, Sitting, Standing, Walking, Running, Cycling, Nordic Walking, Ascending Stairs, Descending Stairs, Vacuum Cleaning, Ironing & 2.54, 9.90, 9.53, 9.78, 12.29, 5.06, 8.47, 9.68, 6.03, 5.41, 9.03, 12.29 \\
\midrule
UCI-HAR & 6 & Walking, Walking Upstairs, Walking Downstairs, Sitting, Standing, Laying & 16.72, 14.99, 13.65, 17.25, 18.51, 18.88 \\
\midrule
USC-HAD & 12 & Walking Forward, Walking Left, Walking Right, Walking Upstairs, Walking Downstairs, Running Forward, Jumping, Sitting, Standing, Sleeping, Elevator Up, Elevator Down & 13.56, 9.42, 9.60, 7.51, 7.04, 6.28, 3.83, 9.31, 8.39, 13.32, 5.86, 5.88 \\
\midrule
Shoaib & 7 & Walking, Standing, Jogging, Sitting, Biking, Walking Upstairs, Walking Downstairs & 14.48, 14.50, 14.50, 14.50, 14.50, 13.05, 14.48 \\
\midrule
MHealth & 13 & Null, Standing still, Sitting and relaxing, Lying down, Walking, Climbing stairs, Waist bends forward, Frontal elevation of arms, Knees bending, Cycling, Jogging, Running, Jump front \& back & 71.73, 2.53, 2.53, 2.53, 2.53, 2.53, 2.32, 2.43, 2.42, 2.53, 2.53, 2.52, 0.86 \\
\midrule
MHealth (no-null) & 12 & Standing still, Sitting and relaxing, Lying down, Walking, Climbing stairs, Waist bends forward, Frontal elevation of arms, Knees bending, Cycling, Jogging, Running, Jump front \& back & 8.88, 8.98, 8.95, 8.92, 8.97, 8.30, 8.56, 8.56, 8.97, 8.94, 8.98, 3.00 \\
\bottomrule
\end{tabularx}
\end{table*}

\paragraph{USC Human Activity Dataset (USC-HAD) \citep{zhang2012usc}.}
USC-HAD consists of six sensor readings from body-worn 3-axis accelerometers and gyroscopes, collected from 14 subjects. The data is sampled at 100~Hz across six channels and includes 12 activity class labels. For evaluation, we follow a leave-subjects-out protocol with subjects 11--12 as the test set and subjects 13--14 as the validation set, while the remaining subjects' data are used for training. A window size $w = 400$ with a stride of 200 (4.0~s, 50\% overlap) is used.

\paragraph{UCI Human Activity Recognition Dataset (UCI-HAR) \citep{anguita2013public}.}
UCI-HAR includes data collected from 30 volunteers performing six activities while wearing a smartphone on their waist. The embedded accelerometer and gyroscope sensors sampled data at 50~Hz, and the pre-processed release provides nine inertial channels (body acceleration, total acceleration, and body gyroscope, each along three axes). The dataset was partitioned into 70\% for training and 30\% for testing. A window size $w = 128$ with a stride of 64 (2.56~s, 50\% overlap) is used.

\paragraph{Physical Activity Monitoring Dataset (PAMAP2) \citep{reiss2012introducing}.}
PAMAP2 includes data from nine subjects wearing IMUs on the chest, the wrist of the dominant arm, and the dominant side's ankle, together with a heart-rate monitor. Each IMU contributes 17 channels (temperature, two 3D accelerometers, gyroscope, magnetometer, and orientation), and after concatenating the three IMUs with the heart-rate channel we obtain 52 input channels covering 12 activity class labels. We follow the standard leave-subjects-out protocol, with the remaining subjects' data used for training. The sample rate is approximately 33~Hz after downsampling, and a window size $w = 24$ with a stride of 12 ($\sim$0.73~s, 50\% overlap) is used.

\paragraph{Mobile Health Dataset (MHealth) \citep{banos2014mhealthdroid}.}
MHealth contains body motion and vital sign recordings from ten volunteers. Sensors were placed on the chest, right lower arm, and left ankle of each subject. For our experiments, we use acceleration and ECG data from the chest, accelerometer, gyroscope, and magnetometer data from the left ankle, and accelerometer, gyroscope, and magnetometer data from the right lower arm, resulting in a total of 23 channels. The data is sampled at 50~Hz and includes 12 activity class labels plus a null class (13 labels in total). Subjects 1--7 are used for training, subject 8 for validation, and subjects 9--10 as the test set. A window size $w = 100$ with a stride of 50 (2.0~s, 50\% overlap) is used.

\paragraph{Shoaib Sensors Activity Recognition Dataset \citep{shoaib2014fusion}.}
This dataset contains smartphone-based inertial recordings from 10 participants performing seven physical activities (walking, standing, jogging, sitting, biking, walking upstairs, and walking downstairs) with phones simultaneously worn at five body positions: left pocket, right pocket, wrist, upper arm, and belt. Each position contributes 9 channels (3-axis accelerometer, gyroscope, and linear accelerometer; the magnetometer is dropped), yielding 45 input channels in total. The data is sampled at 50~Hz. We adopt a leave-two-subjects-out protocol: participants 1 and 9 form the test set, participant 10 the validation set, and participants 2--8 the training set. A window size $w = 100$ with a stride of 50 (2.0~s, 50\% overlap) is used.

\paragraph{CAPTURE-24 \citep{chan2024capture24}.}
CAPTURE-24 is a large-scale dataset featuring 3-channel wrist-worn accelerometer data collected in free-living settings for over 24 hours per participant, with annotations mapped to the 6-class Willetts scheme \citep{willetts2018statistical}: Sleep, Sit-stand, Walking, Bicycling, Mixed, and Vehicle. Given the scale of the full release, all classification experiments use a stratified subset of the data, and LLM-based methods together with \textsc{TRACE-TS} are evaluated on a 25k-sample subset. The sample rate is 100~Hz, and a window size $w = 200$ with a stride of 100 (2.0~s, 50\% overlap) is used.

\paragraph{Handling of Null Classes.}
Two datasets, Opportunity and MHealth, include a Null class in addition to their labelled activities. Because asking a reasoning model to generate a trace justifying the \emph{absence} of an activity is ill-defined and pollutes the training signal, we drop Null-class windows prior to training and evaluation on both datasets. HAR baselines are retrained on the same Null-free data so that both families of models are evaluated on an identical label space. All classification results for Opportunity and MHealth in this paper are therefore reported without the Null class.

\section{Artifact Licensing and Terms of Use}
\label{sec:appendix_licensing}

Table~\ref{tab:licenses} summarizes the licenses of all datasets and pretrained models used in this work. We verified each license from its official distribution: the UCI Machine Learning Repository, the Oxford University Research Archive (ORA), the dataset authors' release pages, and the official Hugging Face / source-repository model cards.

\begin{table}[t]
\caption{Licenses of datasets and core pretrained models used in this work, verified from each artifact's official distribution. $^{\dagger}$Released by the authors as publicly and freely available for research; no formal license file is attached. $^{\ddagger}$Distributed with an open-access article published under CC BY 3.0. $^{\S}$Llama 3.x Community License with its Acceptable Use Policy.}
\label{tab:licenses}
\centering
\small
\setlength{\tabcolsep}{4pt}
\renewcommand{\arraystretch}{1.15}
\begin{tabular}{@{}lll@{}}
\toprule
\textbf{Artifact} & \textbf{License} & \textbf{Source} \\
\midrule
\multicolumn{3}{@{}l}{\textit{Datasets}} \\
UCI-HAR      & CC BY 4.0            & UCI \\
USC-HAD      & Free for research$^{\dagger}$ & USC-SIPI \\
PAMAP2       & CC BY 4.0            & UCI \\
CAPTURE-24   & CC BY 4.0            & Oxford ORA \\
MHEALTH      & CC BY 4.0            & UCI \\
SHOAIB       & CC BY 3.0$^{\ddagger}$ & Univ.\ Twente \\
OPPORTUNITY  & CC BY 4.0            & UCI \\
\midrule
\multicolumn{3}{@{}l}{\textit{Models}} \\
Gemma~4         & Apache 2.0          & Hugging Face \\
Qwen~3.5        & Apache 2.0          & Hugging Face \\
Llama~3.2/3.3   & Llama Comm.\ Lic.$^{\S}$ & Hugging Face \\
GPT-OSS         & Apache 2.0          & Hugging Face \\
MantisV2        & Apache 2.0          & Hugging Face \\
Chronos-2       & Apache 2.0          & Hugging Face \\
\bottomrule
\end{tabular}
\end{table}

\paragraph{Datasets.}
Five of the seven datasets, UCI-HAR, PAMAP2, MHEALTH, and OPPORTUNITY (UCI Machine Learning Repository) together with CAPTURE-24 (Oxford University Research Archive), are released under the Creative Commons Attribution 4.0 International (CC BY 4.0) license, which permits sharing and adaptation for any purpose provided appropriate credit is given. USC-HAD is made publicly and freely available for research by its authors; no formal license file is attached, and we use it solely for non-commercial academic research, consistent with this stated availability. The SHOAIB dataset is released together with an open-access article published under the Creative Commons Attribution (CC BY 3.0) license and is made publicly available for research by its authors. Our use of all datasets, namely training and evaluating activity-recognition and reasoning models, is consistent with their intended research use, and each dataset is credited through citation. All datasets are distributed without personally identifiable information and reference participants only by anonymous subject identifiers.

\paragraph{Models and software.}
Gemma~4, Qwen~3.5 (including the teacher and judge variants), GPT-OSS, MantisV2, and Chronos-2 are released under the permissive Apache~2.0 license. The Llama~3.2/3.3 backbones are used under the Llama~3.x Community License and its Acceptable Use Policy; our research use falls within these terms, and we note that the license requires ``Built with Llama'' attribution for downstream use. The remaining baselines (Attend-\&-Discriminate, NST, PatchTST, UniMTS, SensorLLM, OpenTSLM, and TimeMQA) are open-source research implementations used under their respective licenses. Core software libraries (PyTorch, Hugging Face Transformers and PEFT, vLLM, \texttt{rouge-score}, \texttt{bert-score}, and NLTK) are permissively licensed open-source packages (BSD-, Apache-2.0-, or MIT-style).

\section{Baseline Descriptions}
\label{sec:appendix_baselines}
\subsection{Classification Baselines}
\label{sec:appendix_baselines_classification}
We compare \textsc{TRACE-TS} against several state-of-the-art baselines spanning specialized HAR architectures, time-series foundation models, and general-purpose Transformer-based forecasters. These baselines were selected to provide a comprehensive benchmark across both task-specific and foundation-model paradigms.

\paragraph{Attend \& Discriminate \citep{abedin2021attend}.}
Attend \& Discriminate is a deep learning framework designed specifically for HAR from multi-channel wearable sensor data. It introduces three complementary components: a cross-channel interaction encoder that exploits latent relationships between sensor modalities through self-attention, a data-agnostic augmentation strategy (mixup-style perturbations) to regularize multi-modal sensor streams, and a center-loss-based classification objective that minimizes intra-class variance while maximizing inter-class separation. These design choices yield strong performance across diverse HAR benchmarks, making it a representative specialized HAR baseline.

\paragraph{Mantis V2 \citep{feofanov2026mantisv2}.}
Mantis V2 is a lightweight, calibrated foundation model for time-series classification built on a Vision Transformer (ViT) backbone and pre-trained via self-supervised contrastive learning on synthetic data. It is a refined variant of the original Mantis model, obtained through controlled architectural ablations, producing a more lightweight encoder while substantially narrowing the gap between frozen and fine-tuned performance. It supports self-ensembling and cross-model embedding fusion, and a channel-level adapter enables application to multivariate inputs without independently processing each channel. For baseline, Mantis V2 is fine tuned with a linear layer for classification.

\paragraph{UniMTS \citep{zhang2024unimts}.}
UniMTS is a unified pre-training framework for motion time series that targets generalization across diverse sensor positions, orientations, and activity classes. It employs a contrastive learning objective that aligns motion signals with LLM-enriched text descriptions of activities. To address the scarcity of large-scale motion sensor data, UniMTS synthesizes signals from motion-skeleton datasets using a spatio-temporal graph network, and applies rotation-invariant augmentation to improve robustness across device mounting orientations. For baseline, UniMTS is fine tuned with a linear layer for classification.

\paragraph{Chronos-2 \citep{ansari2025chronos2}.}
Chronos-2 is an encoder-only time-series foundation model (120M parameters) designed for zero-shot forecasting across univariate, multivariate, and covariate-informed tasks within a single architecture. Inspired by the T5 encoder, it introduces a group attention mechanism for efficient in-context learning across related series and covariates, and is trained on a combination of real-world and large-scale synthetic datasets. For baseline, Chronos-2 is used as a frozen encoder with an MLP for classification. (MLP is used here instead of linear layer since the rest of the model is frozen) 

\paragraph{Non-Stationary Transformer (NST) \citep{liu2022non}.}
NST addresses the over-stationarization problem that arises when standard normalization removes the intrinsic non-stationary structure of real-world time series. It introduces two modules: Series Stationarization, which normalizes inputs to improve predictability, and De-stationary Attention, which restores non-stationary information into the attention computation. The framework is model-agnostic and can be plugged into standard Transformer variants. For baseline, NST is trained with a linear layer for classification.

\paragraph{PatchTST \citep{Yuqietal-2023-PatchTST}.}
PatchTST is a Transformer-based model for multivariate time series tasks that uses subseries-level patches as input tokens and a channel-independent approach to reduce computation and improve efficiency. This design retains local semantics, allows for longer historical context, and significantly improves long-term forecasting accuracy. For baseline, PatchTST is trained with a linear layer for classification.

\subsection{Reasoning Baselines}
\label{sec:appendix_baselines_reasoning}

Beyond the classification-oriented baselines above, we compare against a family of LLM-based methods that couple time-series encoders with language models to produce natural-language outputs and reasoning over sensor signals. These methods are the most direct points of comparison for \textsc{TRACE-TS}, as they target language-mediated understanding of time series rather than label-only prediction.

\paragraph{SensorLLM \citep{li2025sensorllm}.}
SensorLLM aligns a frozen large language model with multi-channel motion-sensor data for human activity recognition. It uses a two-stage procedure: a sensor-language alignment stage in which an automatically generated, trend-descriptive textual summary of each channel is paired with the raw signal so the LLM learns to attend to per-channel temporal structure, followed by a task-tuning stage that adapts the aligned model for activity classification. By grounding the LLM in channel-level trend descriptions, SensorLLM closes much of the gap to specialized HAR architectures while retaining a language interface, making it a representative LLM-based HAR baseline.

\paragraph{OpenTSLM \citep{langer2025opentslm}.}
OpenTSLM is a time-series language model that treats continuous time series as a native modality alongside text, enabling reasoning over multivariate signals interleaved with natural-language context. It introduces a soft-prompting formulation in which time series are encoded into learned tokens that are consumed directly by the language model, and the architecture is trained to answer questions and produce explanations grounded in the underlying signal. OpenTSLM is designed for multivariate medical text-and-time-series reasoning including several HAR tasks, and we adapt it to the wearable HAR setting as a state-of-the-art time-series-language reasoning baseline.

\paragraph{Time-MQA \citep{kong-etal-2025-time}.}
Time-MQA is a unified multi-task question-answering framework that casts diverse time-series tasks, spanning numerical analysis and open-ended, reasoning-oriented question answering, as natural-language queries over time series. It is built on the large-scale TSQA dataset (${\sim}200$k question-answer pairs) and continually pre-trains general-purpose LLMs (e.g., Mistral~7B, Llama-3~8B, Qwen-2.5~7B) on this corpus to elicit time-series reasoning that goes beyond purely numeric tasks. We include Time-MQA as a reasoning baseline that probes whether broad QA-style pre-training transfers to grounded activity reasoning on wearable signals.

\section{Implementation Details}
\label{sec:appendix_impl}

\paragraph{Software environment.}
All experiments are implemented in Python~3.11 using PyTorch~2.7 with BF16 mixed-precision training via PyTorch AMP. Models are loaded and fine-tuned with Hugging Face Transformers and PEFT libraries. Flash Attention~2 is used for Qwen and LLaMA backbones; Gemma~4 uses eager attention (see Appendix~\ref{sec:appendix_impl_students}). Teacher inference uses vLLM~0.8.0.

\paragraph{Evaluation packages.}
NLG metrics are computed using the following standard libraries: ROUGE-L via the \texttt{rouge-score} Python package; BERTScore via the \texttt{bert-score} library with \texttt{roberta-large} as the scoring backbone (rescaled against reference); METEOR via the NLTK~3.9 implementation. SNM evaluation uses Qwen3.5-35B-A3B as the primary judge, served via vLLM with the same configuration as the teacher (§\ref{sec:appendix_impl_teacher}); Gemma-4-31B, LLaMA-3.3-70B, and GPT-OSS-120B serve as the additional judges for the four-judge average in Table~\ref{tab:snm_results} and the per-judge analysis in Appendix~\ref{app:snm_judge}.

\subsection{TRACE-TS: Cross-Attention Adapter Training}
\label{sec:appendix_impl_trace}

\textsc{TRACE-TS} keeps the LLM backbone entirely frozen throughout training. Only two components are updated: the \emph{SensorProjector}, which maps MantisV2 embeddings to $N{=}8$ sensor tokens in the LLM's hidden dimension; and the \emph{SensorCrossAttentionAdapters}, one injected at every decoder layer. Each adapter applies gated cross-attention where the LLM hidden states serve as queries and the sensor token sequence serves as keys and values, with a learnable scalar gate initialised at 0.01 to allow smooth warm-up from a text-only regime.

Training minimizes the standard causal language-modeling loss over the teacher-generated structured reasoning traces. No auxiliary classification loss is used; early stopping monitors \texttt{val\_total\_loss} on a held-out validation split. All experiments use identical hyperparameters across the seven benchmarks (Table~\ref{tab:hyperparams}).

\begin{table}[t]
\caption{\textsc{TRACE-TS} training hyperparameters, shared across all backbone variants and datasets.}
\label{tab:hyperparams}
\centering
\small
\begin{tabular}{lc}
\toprule
\textbf{Hyperparameter} & \textbf{Value} \\
\midrule
\multicolumn{2}{l}{\textit{Architecture}} \\
Sensor tokens ($N$) & 8 \\
Adapter rank & 128 \\
Adapter layers & All decoder layers \\
Adapter dropout & 0.1 \\
Gate initialisation & 0.01 \\
\midrule
\multicolumn{2}{l}{\textit{Optimisation}} \\
Optimiser & AdamW \\
Learning rate & $5 \times 10^{-5}$ \\
LR schedule & Cosine + linear warmup \\
Warmup steps & 200 \\
Max epochs & 10 \\
Batch size & 8 \\
Gradient clipping & 1.0 \\
Mixed precision & BF16 (AMP) \\
Early stopping patience & 3 epochs \\
Early stopping $\delta$ & $10^{-4}$ \\
Training parallelism & 8$\times$ H100 80\,GB (DDP) \\
\bottomrule
\end{tabular}
\end{table}

\paragraph{Reporting convention.}
All classification and generation results reported in the main paper and appendix correspond to a single training run per backbone×dataset combination. Given the scale of evaluation (6 backbone variants $\times$ 7 datasets = 42 combinations), multi-seed replication for every setting is computationally prohibitive; we therefore report single-run results and note this wherever results are presented.

\subsection{Teacher Model: Reasoning Generation}
\label{sec:appendix_impl_teacher}

Structured reasoning traces are generated offline using \textbf{Qwen3.5-122B-A10B} as the teacher, served via vLLM with tensor parallelism~8 across eight H100 80\,GB GPUs (\texttt{gpu\_memory\_utilization=0.92}, \texttt{enforce\_eager=False}). Thinking mode is disabled (\texttt{enable\_thinking=False}) to prevent chain-of-thought tokens from polluting the structured output. Generation uses temperature~0.7, top-$p$~0.9, and a maximum of 1024 output tokens per sample.

The teacher receives a prompt containing (i) the per-channel IG and SHAP attribution scores serialised as natural language, (ii) the activity class list for the dataset, and (iii) an instruction to produce a structured DAG in the \textsc{Observation} $\rightarrow$
\textsc{Inference} $\rightarrow$
\textsc{Synthesis} $\rightarrow$
\textsc{Activity} format with explicit \texttt{based\_on} provenance edges. This attribution-conditioned prompting grounds the teacher's reasoning in signals the expert classifier certifiably attends to, rather than spurious correlations.

\subsection{Student Model Variants}
\label{sec:appendix_impl_students}

We evaluate six backbone LLMs spanning three model families. All backbones are loaded in BF16 and kept entirely frozen; only the SensorProjector and cross-attention adapters are trained, with the number of injected adapters equalling the number of decoder layers in each backbone.

\paragraph{Qwen~3.5 (0.8B, 4B)} The Qwen family uses Flash Attention~2 with the standard BF16 compute path. The 0.8B variant has 24 decoder layers; the 4B variant has 32. Both are trained with \texttt{max\_seq\_len}~=~768 and \texttt{max\_new\_tokens}~=~768. Per-GPU batch sizes are 16 (0.8B),
8 (4B),
reduced for larger variants to stay within the 80\,GB H100 VRAM budget.

\paragraph{Llama~3.2 (1B, 3B)} Llama backbones also use Flash Attention~2. The 1B variant has 16 decoder layers and supports a batch size of 32 due to its compact hidden dimension; the 3B has 28 layers and uses a batch size of 16. Both use \texttt{max\_seq\_len}~=~768 and \texttt{max\_new\_tokens}~=~768.

\paragraph{Gemma~4 (E2B-it, E4B-it).} Gemma~4 models are sparse mixture-of-experts architectures with 35 (E2B) and 42 (E4B) decoder layers, and use \emph{eager attention} rather than Flash Attention~2, because their global attention head dimension (512) exceeds Flash Attention~2's per-head dimension cap of 256. Per-GPU batch sizes are 16 (E2B) and 8 (E4B). \texttt{torch.compile} is disabled for all backbones to avoid CUDA graph overwrite errors during distributed training.

\section{Full Classification Results}
\label{app:appendix_full_results}

Table~\ref{tab:full_results} extends Table~\ref{tab:cot_har_ours_comparison} with all \textsc{TRACE-TS} backbone variants and the remaining CoT baselines omitted from the main paper for brevity.

\begin{table*}[t]
\caption{Additional \textsc{TRACE-TS} backbones and baselines for the HAR task. All values in (\%), with the best results in bold and the second-best underlined.}
\label{tab:full_results}
\centering
\resizebox{\textwidth}{!}{%
\begin{tabular}{l|cc|cc|cc|cc|cc|cc|cc|cc}
\toprule
& \multicolumn{2}{c|}{UCI-HAR}
& \multicolumn{2}{c|}{USC-HAD}
& \multicolumn{2}{c|}{PAMAP2}
& \multicolumn{2}{c|}{CAPTURE-24}
& \multicolumn{2}{c|}{MHEALTH}
& \multicolumn{2}{c|}{SHOAIB}
& \multicolumn{2}{c|}{OPPORTUNITY}
& \multicolumn{2}{c}{Average} \\
Method & Acc & F1 & Acc & F1 & Acc & F1 & Acc & F1 & Acc & F1 & Acc & F1 & Acc & F1 & Acc & F1 \\
\midrule
\multicolumn{17}{l}{\textbf{LLM-Based Methods (additional)}} \\
\midrule
OpenTSLM (Zero-shot) & 35.32 & 21.95 & 9.59 & 2.52 & 8.94 & 4.20 & 4.70 & 4.87 & 1.00 & 1.18 & 21.40 & 13.26 & 0.00 & 0.00 & 11.56 & 6.85 \\
Llama 3.2 3B (CoT) & 16.36 & 10.40 & 8.49 & 4.66 & 8.02 & 5.32 & 10.34 & 5.92 & 9.02 & 3.71 & 10.59 & 7.69 & 8.35 & 3.55 & 10.17 & 5.89 \\
Qwen 3.5 4B (CoT)  & 27.70 & 24.24 & 16.78 & 11.06 & 11.50 &  9.42 & 46.14 & 27.49 & 23.02 & 18.63 & 24.96 & 22.75 &  9.02 &  5.48 & 22.73 & 17.01 \\
Qwen 3.5 9B (CoT)  & 29.74 & 26.14 & 15.42 & 12.85 & 13.69 & 11.07 & 46.37 & 26.24 & 23.17 & 19.19 & 28.65 & 24.43 &  6.29 &  3.66 & 23.33 & 17.65 \\
Gemma 4 4B (CoT)   & 21.49 & 13.09 &  9.73 &  3.55 & 10.75 &  3.74 & 21.05 & 15.72 & 12.25 &  3.45 & 13.78 &  5.86 &  4.96 &  3.03 & 13.43 &  6.92 \\
TimeMQA (LLaMA3) & 25.28 & 17.15 & 10.47 & 4.97 & 10.28 & 5.30 & 13.90 & 5.87 & 9.33 & 1.42 & 15.19 & 6.10 & 6.63 & 0.75 & 13.01 & 5.94 \\
TimeMQA (Mistral) & 1.78 & 2.96 & 4.27 & 3.50 & 7.64 & 1.62 & 1.94 & 1.28 & 10.73 & 3.46 & 12.07 & 7.13 & 6.80 & 1.10 & 6.46 & 3.01 \\
LLaSA-13B & 17.13 & 4.88 & 10.19 & 1.54 & 10.25 & 1.55 & 2.13 & 0.70 & 9.33 & 1.42 & 9.28 & 2.43 & 14.13 & 1.46 & 10.35 & 2.00 \\
\midrule
\multicolumn{17}{l}{\textbf{Ours (\textsc{TRACE-TS}): additional backbones}} \\
\midrule
Gemma 4 2B & \textbf{94.60} & \textbf{94.74} & \underline{56.14} & \underline{49.00} & 84.38 & 80.29 & \textbf{63.87} & \textbf{59.87} & 93.04 & 93.24 & \underline{99.17} & \underline{99.17} & 68.56 & 58.31 & \underline{79.97} & \underline{76.37} \\
Llama 3.2 1B & 92.16 & 92.20 & 50.91 & 44.97 & \underline{84.47} & \textbf{82.06} & 62.44 & 56.10 & \underline{97.44} & \underline{97.62} & \underline{99.17} & 99.16 & \textbf{69.28} & \textbf{60.50} & 79.41 & 76.09 \\
Qwen 3.5 0.8B & \underline{92.98} & \underline{93.08} & \textbf{69.47} & \textbf{64.99} & \textbf{84.90} & \underline{81.53} & \underline{63.65} & \underline{58.34} & \textbf{97.95} & \textbf{98.11} & \textbf{99.21} & \textbf{99.20} & \underline{69.16} & \underline{59.94} & \textbf{82.47} & \textbf{79.31} \\
\bottomrule
\end{tabular}%
}
\end{table*}

\paragraph{Opportunity.}
Opportunity is the most challenging benchmark for all methods. Its 17 non-null classes include closely related fine-grained gestures (e.g.\ \textit{Open Drawer 1} vs.\ \textit{Open Drawer 2}, \textit{Close Door 1} vs.\ \textit{Close Door 2}) that differ only subtly in movement, and its class distribution is highly skewed, ranging from 2.43\% to 22.66\% of samples (Table~\ref{tab:dataset_proportions}). Both factors penalize macro-F1 heavily. Specialist HAR models, which are trained purely for discrimination, retain an advantage here: \textsc{TRACE-TS} (Gemma~4 4B) reaches 77.13\% accuracy and 66.63 macro-F1, against the specialist best of 84.33\% / 77.77.

\paragraph{USC-HAD}
On USC-HAD, the leave-subjects-out protocol introduces substantial inter-subject variability across 12 near-uniformly distributed classes. The best \textsc{TRACE-TS} variant (Qwen~3.5 4B, 72.35\% accuracy) is close to the strongest specialist (Mantis~V2, 73.09\%).

\paragraph{MHealth, Shoaib, and CAPTURE-24.}
\textsc{TRACE-TS} performs strongly on the better-separated benchmarks. All backbone variants exceed 93\% accuracy on Shoaib and 92\% on MHealth, where activities such as jogging, cycling, and standing produce clearly distinguishable signals. CAPTURE-24 is harder despite its small label space, as its three wrist-worn accelerometer channels and free-living categories (Sleep, Mixed, Vehicle) are difficult to separate within a short window; \textsc{TRACE-TS} (Gemma~4 4B) nonetheless reaches 63.96\% accuracy, competitive with the strongest baselines on this dataset.

\paragraph{Zero-shot and CoT baselines.}
Table~\ref{tab:full_results} also shows that LLM-based methods without learned sensor conditioning perform close to chance across every benchmark. Zero-shot OpenTSLM averages 11.56\% accuracy and 6.85 macro-F1, the CoT-prompted variants of Llama~3.2 3B, Qwen~3.5 4B, and Qwen~3.5 9B average between 10.17\% and 23.33\% accuracy, and both TimeMQA variants are comparable or lower. By contrast, every \textsc{TRACE-TS} backbone, including the smallest 0.8B and 1B students, exceeds 79\% average accuracy. This gap indicates that prompting or text-serialising sensor windows is insufficient on its own, and that the learned cross-attention conditioning is what enables competitive recognition.

\section{Generation Quality: All Backbones}
\label{sec:appendix_generation}

Table~\ref{tab:cot_vs_ours_generation} reports BERTScore and METEOR for all \textsc{TRACE-TS} backbone variants and CoT baselines across seven benchmarks. These scores complement the correctness-conditioned analysis in Table~\ref{tab:cot_conditioned_generation}.

\begin{table*}[t]
\caption{Generation quality (BERTScore and METEOR) for all \textsc{TRACE-TS} backbone variants and CoT baselines. All values reported in (\%). Best in bold, second-best underlined.} 
\label{tab:cot_vs_ours_generation}
\centering
\scriptsize
\setlength{\tabcolsep}{1.7pt}
\renewcommand{\arraystretch}{0.95}
\begin{tabular}{l|cc|cc|cc|cc|cc|cc|cc|cc}
\toprule
& \multicolumn{2}{c|}{UCI-HAR} & \multicolumn{2}{c|}{USC-HAD} & \multicolumn{2}{c|}{PAMAP2} & \multicolumn{2}{c|}{CAPTURE-24} & \multicolumn{2}{c|}{MHEALTH} & \multicolumn{2}{c|}{SHOAIB} & \multicolumn{2}{c|}{OPPORTUNITY} & \multicolumn{2}{c}{Average} \\
Model & BERTSc & MET. & BERTSc & MET. & BERTSc & MET. & BERTSc & MET. & BERTSc & MET. & BERTSc & MET. & BERTSc & MET. & BERTSc & MET. \\
\midrule
\multicolumn{17}{l}{\textbf{CoT Baselines}} \\
\midrule
Llama 3.2 3B & 90.88 & 28.30 & 90.83 & 27.87 & 90.79 & 25.73 & 89.80 & 24.80 & 91.04 & 29.15 & 90.67 & 28.38 & 90.74 & 28.58 & 90.68 & 27.54 \\
Llama 3.3 70B & 91.06 & 26.17 & 91.36 & 28.81 & 90.61 & 23.11 & 89.53 & 21.63 & 90.45 & 23.98 & 90.76 & 26.59 & 90.16 & 23.83 & 90.56 & 24.87 \\
Qwen 3.5 4B  & 90.40 & 29.39 & 90.09 & 27.15 & 90.39 & 28.36 & 89.11 & 25.44 & 90.10 & 30.09 & 90.07 & 29.77 & 90.64 & 31.73 & 90.11 & 28.85 \\
Qwen 3.5 9B  & 90.61 & 30.23 & 90.27 & 27.51 & 90.76 & 28.66 & 89.23 & 25.37 & 90.19 & 29.06 & 89.91 & 28.25 & 90.59 & 30.23 & 90.22 & 28.47 \\
Qwen 3.5 27B & 90.38 & 30.20 & 90.27 & 27.88 & 90.65 & 29.04 & 89.08 & 25.71 & 90.20 & 31.66 & 90.19 & 31.94 & 90.54 & 31.98 & 90.19 & 29.77 \\
Gemma 4 31B  & 90.85 & 31.83 & 90.76 & 29.30 & 91.26 & 30.49 & 89.54 & 26.02 & 90.68 & 31.60 & 90.91 & 32.88 & 90.83 & 32.19 & 90.69 & 30.62 \\
Gemma 4 4B   & 91.04 & 32.08 & 91.16 & 31.98 & 91.09 & 32.48 & 89.89 & 26.89 & 90.66 & 33.53 & 90.92 & 33.77 & 91.13 & 34.29 & 90.84 & 32.15 \\
GPT-OSS 120B & 90.02 & 26.70 & 89.56 & 24.39 & 90.46 & 27.31 & 88.98 & 23.73 & 89.94 & 27.15 & 90.20 & 28.77 & 90.41 & 29.25 & 89.94 & 26.76 \\
\midrule
\multicolumn{17}{l}{\textbf{Ours (\textsc{TRACE-TS})}} \\
\midrule
Gemma 4 2B & 95.85 & 61.59 & \underline{95.87} & 60.37 & \underline{95.28} & 59.62 & 95.96 & \underline{62.33} & 95.35 & 58.61 & 95.62 & 59.65 & 95.31 & 62.31 & 95.61 & 60.64 \\
Gemma 4 4B & 95.88 & \textbf{61.71} & \textbf{95.92} & \textbf{60.76} & \textbf{95.30} & \textbf{60.49} & \underline{95.97} & \textbf{62.72} & 95.35 & 58.52 & \underline{95.66} & \textbf{60.28} & \underline{95.40} & \textbf{62.59} & \underline{95.64} & \textbf{61.01} \\
Llama 3.2 1B & 95.88 & \underline{61.68} & 95.86 & 60.48 & 95.21 & 60.12 & 95.95 & 62.04 & 95.48 & \textbf{59.26} & 95.63 & 59.82 & \textbf{95.42} & \underline{62.44} & 95.63 & \underline{60.83} \\
Llama 3.2 3B & \underline{95.89} & 61.67 & 95.85 & 60.46 & 95.22 & \underline{60.22} & \underline{95.97} & 62.26 & \textbf{95.50} & \underline{58.81} & 95.59 & 59.23 & 95.31 & 62.19 & 95.62 & 60.69 \\
Qwen 3.5 0.8B & \textbf{95.92} & 58.84 & \underline{95.87} & 59.35 & 95.25 & 59.31 & \textbf{95.98} & 59.97 & \underline{95.49} & 55.63 & \textbf{95.71} & 57.61 & \textbf{95.42} & 58.58 & \textbf{95.66} & 58.47 \\
Qwen 3.5 4B & 95.86 & 52.95 & 95.80 & 56.22 & 95.16 & 57.94 & 95.96 & 60.80 & 95.29 & 58.27 & 95.51 & 59.00 & 95.23 & 59.66 & 95.54 & 57.83 \\
\bottomrule
\end{tabular}
\end{table*}

\paragraph{METEOR}
Every \textsc{TRACE-TS} variant scores substantially higher on METEOR than any CoT baseline, with a gap of roughly 30 points on each dataset. As METEOR rewards unigram precision, recall, and low fragmentation, this gap reflects that \textsc{TRACE-TS} traces reproduce the teacher's terminology and node structure, whereas CoT outputs are free-form and draw on language priors rather than the attribution evidence available to the teacher.

\paragraph{BERTScore.}
BERTScore is far less discriminative. All scores fall in a narrow band (roughly 89--96\%), with \textsc{TRACE-TS} variants clustered near the top (95--96\%) and the CoT baselines only a few points lower (89--91\%). The compression arises because the pretrained encoder assigns high similarity to topically related sentences even when they describe different signals. This is consistent with Table~\ref{tab:cot_conditioned_generation}, where correctly and incorrectly classified samples differ by at most 0.56 BERTScore points, motivating the SNM metrics as the primary measure of reasoning fidelity.

\paragraph{Backbone differences.}
Within \textsc{TRACE-TS}, Gemma~4 4B obtains the best or second-best METEOR on six of the seven datasets, while Qwen~3.5 0.8B obtains the highest BERTScore on several. This dissociation reflects what each metric measures: BERTScore favours the dense, topically coherent text of the smaller model, while METEOR rewards the larger model's closer reproduction of the teacher's vocabulary and structure. A model strong on both therefore produces traces that are simultaneously semantically aligned and structurally faithful.

\section{SNM Across Methods: Per-Dataset Results}
\label{app:snm_all_methods}

Table~\ref{tab:snm_all_methods} reports per-dataset SNM for representative \textsc{TRACE-TS} backbones alongside the external CoT baselines and the fine-tuned OpenTSLM baseline. Both \textsc{TRACE-TS} backbones recover reference-level fidelity on all three SNM axes; zero-shot CoT baselines collapse on observation matching, with average SNM-OF1 below 1\% for every CoT model (and never above 2.2\% on any single dataset); the fine-tuned OpenTSLM baseline sits between the two regimes.

\begin{table*}[t]
\caption{Per-dataset SNM-OF1, SNM-IF1, and SNM-SA (\%) for representative \textsc{TRACE-TS} backbones, external CoT baselines, and the fine-tuned OpenTSLM baseline.}
\label{tab:snm_all_methods}
\centering
\scriptsize
\setlength{\tabcolsep}{1.7pt}
\renewcommand{\arraystretch}{0.95}
\resizebox{\textwidth}{!}{%
\begin{tabular}{l|ccc|ccc|ccc|ccc|ccc|ccc|ccc|ccc}
\toprule
& \multicolumn{3}{c|}{UCI-HAR} & \multicolumn{3}{c|}{USC-HAD} & \multicolumn{3}{c|}{PAMAP2} & \multicolumn{3}{c|}{CAPTURE-24} & \multicolumn{3}{c|}{MHEALTH} & \multicolumn{3}{c|}{SHOAIB} & \multicolumn{3}{c|}{OPPORTUNITY} & \multicolumn{3}{c}{Average} \\
Model & OF1 & IF1 & SA & OF1 & IF1 & SA & OF1 & IF1 & SA & OF1 & IF1 & SA & OF1 & IF1 & SA & OF1 & IF1 & SA & OF1 & IF1 & SA & OF1 & IF1 & SA \\
\midrule
\multicolumn{25}{l}{\textbf{CoT Baselines}} \\
\midrule
Gemma 4 4B (E4B)  &  0.23 & 12.93 & 13.09 &  0.15 & 10.74 &  5.22 &  0.20 &  7.97 &  4.45 &  0.00 & 13.30 &  6.23 &  0.27 &  9.54 &  6.53 &  0.19 & 13.67 &  6.96 &  0.01 &  2.61 &  3.45 &  0.15 & 10.11 &  6.56 \\
Qwen 3.5 27B      &  0.26 & 37.48 & 32.24 &  0.29 &  9.54 &  3.87 &  0.34 & 18.02 & 12.76 &  0.00 & 26.22 & 30.97 &  0.84 & 21.07 & 29.47 &  0.42 & 30.08 & 30.56 &  0.09 &  5.00 &  4.48 &  0.32 & 21.06 & 20.62 \\
Gemma 4 31B       &  0.38 & 33.32 & 36.32 &  0.44 & 14.19 & 13.55 &  0.31 & 17.14 & 16.15 &  0.00 & 20.81 & 28.82 &  1.16 & 24.78 & 34.06 &  0.44 & 33.80 & 30.38 &  0.11 &  6.83 &  9.47 &  0.41 & 21.55 & 24.11 \\
LLaMA 3.3 70B     &  2.19 & 21.23 & 16.71 &  0.62 & 11.92 &  8.36 &  0.57 & 12.18 &  4.41 &  0.00 & 29.76 & 11.93 &  0.68 & 10.75 &  4.98 &  0.69 & 24.24 & 22.24 &  0.02 &  5.02 &  1.21 &  0.68 & 16.44 &  9.98 \\
GPT-OSS 120B      &  1.53 & 31.95 & 31.16 &  0.39 &  7.90 &  5.33 &  0.71 & 14.81 & 13.12 &  0.00 & 30.04 & 45.48 &  0.80 & 21.68 & 26.59 &  1.09 & 26.35 & 25.78 &  0.17 &  6.42 &  6.29 &  0.67 & 19.88 & 21.97 \\
\midrule
\multicolumn{25}{l}{\textbf{Fine-tuned Baseline}} \\
\midrule
OpenTSLM          & 15.95 & 23.25 & 17.93 &  8.77 &  6.84 &  4.56 &  6.16 & 12.30 &  8.19 & 15.31 & 22.36 & 27.18 &  5.88 &  7.08 &  5.99 &  5.31 & 12.20 & 15.49 &  9.09 &  7.10 & 10.08 &  9.50 & 13.02 & 12.77 \\
\midrule
\multicolumn{25}{l}{\textbf{Ours (\textsc{TRACE-TS})}} \\
\midrule
Gemma 4 4B        & 32.58 & 53.42 & 92.51 & 25.30 & 40.12 & 63.67 &  9.37 & 26.28 & 60.53 & 30.68 & 53.03 & 82.60 & 21.89 & 38.35 & 79.24 & 19.89 & 46.94 & 93.08 & 10.74 & 15.98 & 61.33 & 21.49 & 39.16 & 76.14 \\
Llama 3.2 3B      & 33.48 & 52.74 & 90.11 & 29.24 & 39.24 & 61.89 &  9.95 & 26.32 & 57.28 & 31.38 & 53.75 & 78.98 & 23.54 & 41.93 & 80.48 & 21.10 & 47.09 & 92.36 & 11.94 & 17.98 & 61.24 & 22.95 & 39.87 & 74.62 \\
\bottomrule
\end{tabular}%
}
\end{table*}

\section{SNM Judge Robustness}
\label{app:snm_judge}

Table~\ref{tab:snm_per_judge} reports SNM-OF1, SNM-IF1, and SNM-SA for each of the four LLM judges. Gemma-4-31B and LLaMA-3.3-70B yield consistently higher scores than Qwen3.5-35B and GPT-OSS-120B, particularly on SNM-IF1, reflecting differences in judge calibration and instruction-following strictness. Despite this spread, the rank ordering across datasets is consistent, confirming that the conclusions drawn from SNM are not artefacts of a single judge.

\begin{table*}[t]
\caption{Per-dataset SNM-OF1, SNM-IF1, and SNM-SA (\%) for \textsc{TRACE-TS} (Gemma~4 4B) evaluated with each of the four LLM judges. }
\label{tab:snm_per_judge}
\centering
\small
\resizebox{\textwidth}{!}{%
\begin{tabular}{l|cccc|cccc|cccc}
\toprule
& \multicolumn{4}{c|}{\textbf{SNM-OF1} (\%)} & \multicolumn{4}{c|}{\textbf{SNM-IF1} (\%)} & \multicolumn{4}{c}{\textbf{SNM-SA} (\%)} \\
Dataset & Qwen 35B & Gemma 31B & LLaMA 70B & GPT-OSS 120B & Qwen 35B & Gemma 31B & LLaMA 70B & GPT-OSS 120B & Qwen 35B & Gemma 31B & LLaMA 70B & GPT-OSS 120B \\
\midrule
UCI-HAR      & 32.58 & 48.36 & 51.24 & 29.32 & 53.42 & 88.70 & 89.76 & 51.77 & 92.51 & 96.83 & 97.11 & 74.07 \\
USC-HAD      & 25.30 & 42.99 & 43.29 & 30.33 & 40.12 & 77.88 & 79.87 & 45.15 & 63.67 & 71.51 & 74.44 & 53.55 \\
PAMAP2       &  9.37 & 20.13 & 19.82 & 12.02 & 26.28 & 67.40 & 71.45 & 41.12 & 60.53 & 70.85 & 71.06 & 55.53 \\
CAPTURE-24   & 30.68 & 52.11 & 52.68 & 32.70 & 53.03 & 81.56 & 81.76 & 47.35 & 82.60 & 87.95 & 88.39 & 61.83 \\
MHEALTH      & 21.89 & 41.62 & 42.30 & 26.85 & 38.35 & 78.32 & 80.33 & 56.00 & 79.24 & 97.98 & 97.43 & 79.95 \\
SHOAIB       & 19.89 & 34.81 & 35.49 & 15.72 & 46.94 & 86.03 & 87.47 & 44.04 & 93.08 & 99.20 & 99.41 & 86.67 \\
OPPORTUNITY  & 10.74 & 24.06 & 24.20 & 18.39 & 15.98 & 67.79 & 73.05 & 47.71 & 61.33 & 81.40 & 80.28 & 73.90 \\
\bottomrule
\end{tabular}%
}
\end{table*}

\section{Reasoning Perturbation: Per-Dataset Results}
\label{app:perturbation_ablation}

Table~\ref{tab:perturbation_ablation_full} reports per-dataset BERTScore, METEOR, and SNM metrics for the reference-trace perturbation ablation in Section~\ref{sec:ablation}. \textit{Paraphrase} rewrites clauses while preserving sensor claims and provenance; \textit{Misgrounded} rewires cited channels and \texttt{based\_on} identifiers with plausible-looking wrong ones while leaving surface text intact.

\begin{table*}[t]
\caption{Per-dataset perturbation ablation on reference reasoning traces. All values in (\%). Orig / Paraph / Misgr = unperturbed, paraphrased, and misgrounded references. $^{\dagger}$Macro-F1 is unchanged: perturbation affects only the reference traces, not model predictions.}
\label{tab:perturbation_ablation_full}
\centering\small
\resizebox{\textwidth}{!}{%
\begin{tabular}{l|c|ccc|ccc|ccc|ccc|ccc}
\toprule
& & \multicolumn{3}{c|}{BERTScore} & \multicolumn{3}{c|}{METEOR} & \multicolumn{3}{c|}{SNM-OF1} & \multicolumn{3}{c|}{SNM-IF1} & \multicolumn{3}{c}{SNM-SA} \\
Dataset & Macro-F1$^{\dagger}$ & Orig & Paraph & Misgr & Orig & Paraph & Misgr & Orig & Paraph & Misgr & Orig & Paraph & Misgr & Orig & Paraph & Misgr \\
\midrule
UCI-HAR      & 96.73 & 95.88 & 95.43 & 95.58 & 61.71 & 56.90 & 55.48 & 32.58 & 32.85 & 33.77 & 53.42 & 52.74 & 28.85 & 92.51 & 91.75 & 10.24 \\
USC-HAD      & 64.36 & 95.92 & 95.41 & 95.44 & 60.76 & 57.07 & 55.84 & 25.30 & 27.15 & 26.94 & 40.12 & 39.45 & 16.66 & 63.67 & 62.89 &  3.87 \\
PAMAP2       & 83.08 & 95.30 & 94.90 & 95.03 & 60.49 & 56.09 & 54.46 &  9.37 &  9.82 &  9.88 & 26.28 & 24.77 & 10.82 & 60.53 & 59.97 &  2.86 \\
CAPTURE-24   & 60.20 & 95.97 & 95.62 & 95.65 & 62.72 & 57.94 & 56.57 & 30.68 & 32.44 & 32.44 & 53.03 & 49.75 & 21.81 & 82.60 & 80.50 &  4.17 \\
MHEALTH      & 98.27 & 95.35 & 94.90 & 95.05 & 58.52 & 53.71 & 51.97 & 21.89 & 23.39 & 23.26 & 38.35 & 36.36 & 10.21 & 79.24 & 78.62 &  2.64 \\
SHOAIB       & 99.40 & 95.66 & 95.26 & 95.39 & 60.28 & 55.71 & 54.05 & 19.89 & 20.12 & 20.76 & 46.94 & 47.40 & 13.62 & 93.08 & 93.46 &  5.11 \\
OPPORTUNITY  & 66.63 & 95.40 & 94.06 & 93.91 & 62.59 & 57.26 & 56.16 & 10.74 & 11.76 & 11.62 & 15.98 & 14.68 &  1.72 & 61.33 & 59.35 &  0.00 \\
\midrule
Average      & 81.24 & 95.64 & 95.08 & 95.15 & 61.01 & 56.38 & 54.93 & 21.49 & 22.50 & 22.67 & 39.16 & 37.88 & 14.81 & 76.14 & 75.22 &  4.13 \\
\bottomrule
\end{tabular}%
}
\end{table*}

\section{Extended Ablation Results}

\subsection{Teacher ablation: Per-Dataset Results}
\label{sec:appendix_llama_teacher}
Table~\ref{tab:llama_teacher_results} reports dataset-level classification results for Gemma~4 4B trained with LLaMA~3.3-70B-Instruct, Gemma~4-31B, GPT-OSS-120B, and the default Qwen~3.5-122B teacher across all seven benchmarks. The consistent ranking (Qwen~122B $>$ GPT-OSS~120B $>$ Gemma~31B $\approx$ LLaMA~70B) demonstrates that stronger teachers provide more effective supervision.

\paragraph{Teacher-sensitive datasets.}
Opportunity, USC-HAD, and MHealth exhibit the largest performance gaps. LLaMA~70B reduces accuracy by 7.00, 5.84, and 5.79 points, respectively, while Gemma~31B incurs a 10.27-point drop on USC-HAD. These datasets contain larger or more fine-grained activity spaces, making the quality of teacher-generated reasoning particularly important. GPT-OSS~120B substantially narrows the gaps (4.05, 2.23, and 3.22), consistent with its stronger reasoning ability.

\paragraph{Teacher-insensitive datasets.}
CAPTURE-24 is largely insensitive to teacher choice (within $\pm0.55$ accuracy points across all teachers), suggesting that its coarse activity taxonomy requires less sophisticated reasoning. Shoaib and UCI-HAR show moderate sensitivity, with LLaMA~70B reducing accuracy by 4.40 and 4.64 points, compared to only 0.99 and 2.17 for GPT-OSS~120B. Overall, teacher quality has the greatest impact on datasets with larger or finer-grained activity spaces.

\begin{table}[t]
\caption{Per-dataset classification results for Gemma~4 4B \textsc{TRACE-TS} using LLaMA~3.3-70B, Gemma~4-31B, GPT-OSS-120B, and Qwen~3.5-122B teachers. All values in (\%); all other settings identical.}
\label{tab:llama_teacher_results}
\centering
\tiny
\resizebox{\columnwidth}{!}{%
\begin{tabular}{l|cc|cc|cc|cc}
\toprule
& \multicolumn{2}{c|}{LLaMA 3.3-70B} & \multicolumn{2}{c|}{Gemma 4-31B} & \multicolumn{2}{c|}{GPT-OSS-120B} & \multicolumn{2}{c}{Qwen 3.5-122B} \\
Dataset & Acc & F1 & Acc & F1 & Acc & F1 & Acc & F1 \\
\midrule
UCI-HAR      & 92.03 & 92.08 & 91.52 & 91.55 & 94.50 & 94.56 & 96.67 & 96.73 \\
USC-HAD      & 62.74 & 63.67 & 58.31 & 60.61 & 66.35 & 64.29 & 68.58 & 64.36 \\
PAMAP2       & 82.31 & 75.65 & 84.13 & 79.79 & 83.88 & 81.92 & 87.10 & 83.08 \\
CAPTURE-24   & 63.86 & 58.84 & 64.01 & 60.23 & 63.41 & 59.72 & 63.96 & 60.20 \\
MHEALTH      & 92.38 & 92.68 & 90.99 & 90.60 & 94.95 & 95.15 & 98.17 & 98.27 \\
SHOAIB       & 95.00 & 94.89 & 97.78 & 97.77 & 98.41 & 98.41 & 99.40 & 99.40 \\
OPPORTUNITY  & 70.13 & 59.93 & 71.70 & 61.25 & 73.08 & 63.26 & 77.13 & 66.63 \\
\midrule
Average      & 79.78 & 76.82 & 79.78 & 77.40 & 82.08 & 79.62 & 84.43 & 81.24 \\
\bottomrule
\end{tabular}%
}
\end{table}

\subsection{Attribution Source Ablation: Per-Dataset Results}
\label{app:attr_ablation}

Table~\ref{tab:attr_ablation_full} reports per-dataset classification results for the attribution source ablation described in Section~\ref{sec:ablation}. All variants use Gemma~4 4B with the same cross-attention adapter configuration as \textsc{TRACE-TS}; only the teacher's attribution signal differs.

Table~\ref{tab:human_ab} reports per-dataset results for the blind pairwise human evaluation described in Section~\ref{sec:ablation}. For each evaluated window, five annotators independently compared the trace generated with the fused IG+SHAP attribution against the trace from the no-attribution variant, blind to which variant produced each trace and with randomized A/B order, and selected the preferred trace or declared a tie (250 judgments in total).

\begin{table}[t]
\caption{Blind pairwise human evaluation between attribution-grounded and no-attribution reasoning traces, aggregated over five annotators. \textit{Win rate} counts only decisive (non-tie) judgments.}
\label{tab:human_ab}
\centering
\small
\resizebox{\columnwidth}{!}{%
\begin{tabular}{lcccc}
\toprule
Dataset & Attr.\ wins (\%) & Ties (\%) & No-attr.\ wins (\%) & Win rate (excl.\ ties, \%) \\
\midrule
Opportunity & 93.3  & 3.3  & 3.3  & 96.6 \\
MHEALTH     & 100.0 & 0.0  & 0.0  & 100.0 \\
PAMAP2      & 83.0  & 9.0  & 8.0  & 91.2 \\
USC-HAD     & 76.0  & 20.0 & 4.0  & 95.0 \\
Shoaib      & 95.0  & 5.0  & 0.0  & 100.0 \\
Capture-24  & 66.7  & 17.8 & 15.6 & 81.1 \\
UCI-HAR     & 85.0  & 0.0  & 15.0 & 85.0 \\
\midrule
All         & 82.4  & 9.6  & 8.0  & 91.2 \\
\bottomrule
\end{tabular}%
}
\end{table}

\begin{table*}[t]
\caption{Per-dataset classification results for the attribution-source ablation. All values in (\%); $\Delta$ denotes change relative to \textsc{TRACE-TS}.}
\label{tab:attr_ablation_full}
\centering\small
\resizebox{\textwidth}{!}{%
\begin{tabular}{l|cc|cc|cc|cc|cc|cc|cc|cccc}
\toprule
 & \multicolumn{2}{c|}{UCI-HAR} & \multicolumn{2}{c|}{USC-HAD}
 & \multicolumn{2}{c|}{PAMAP2} & \multicolumn{2}{c|}{CAPTURE-24}
 & \multicolumn{2}{c|}{MHEALTH} & \multicolumn{2}{c|}{SHOAIB}
 & \multicolumn{2}{c|}{OPPORTUNITY}
 & \multicolumn{4}{c}{Average (7 ds)} \\
Method & Acc & F1 & Acc & F1 & Acc & F1 & Acc & F1 & Acc & F1 & Acc & F1 & Acc & F1 & Acc & F1 & $\Delta$Acc & $\Delta$F1 \\
\midrule
SHAP-only attr.      & 96.06 & 96.14 & 63.46 & 67.17 & 86.38 & 83.03 & 63.10 & 58.76 & 95.46 & 95.63 & 98.89 & 98.89 & 75.02 & 66.33 & 82.62 & 80.85 & $-$1.81 & $-$0.39 \\
IG-only attr.        & 95.96 & 96.00 & 59.32 & 64.47 & 86.12 & 81.00 & 63.65 & 59.67 & 94.65 & 94.77 & 98.73 & 98.73 & 72.78 & 62.03 & 81.60 & 79.52 & $-$2.83 & $-$1.72 \\
Random attr.         & 93.86 & 93.94 & 60.90 & 62.80 & 85.37 & 83.24 & 63.80 & 60.01 & 97.66 & 97.81 & 99.21 & 99.20 & 74.59 & 65.29 & 82.20 & 80.33 & $-$2.23 & $-$0.91 \\
No attribution       & 94.16 & 94.35 & 64.40 & 64.79 & 86.58 & 82.54 & 62.49 & 58.29 & 87.99 & 88.56 & 96.78 & 96.76 & 72.48 & 62.57 & 80.70 & 78.27 & $-$3.73 & $-$2.97 \\
Bottom-10\% attr.    & 94.13 & 94.26 & 56.47 & 64.20 & 85.03 & 79.91 & 62.11 & 57.15 & 90.18 & 90.31 & 98.41 & 98.41 & 73.02 & 61.54 & 79.91 & 77.97 & $-$4.52 & $-$3.27 \\
\midrule
\textsc{TRACE-TS} (default) & 96.67 & 96.73 & 68.58 & 64.36 & 87.10 & 83.08 & 63.96 & 60.20 & 98.17 & 98.27 & 99.40 & 99.40 & 77.13 & 66.63 & 84.43 & 81.24 & --- & --- \\
\bottomrule
\end{tabular}%
}
\end{table*}

\subsection{Output Format Ablation: Per-Dataset Results}

\label{app:output_ablation}

Table~\ref{tab:output_ablation_full} reports per-dataset classification results for the three output format ablation variants described in Section~\ref{sec:ablation}. All variants use the same Gemma~4 4B architecture and cross-attention adapter configuration as \textsc{TRACE-TS} (full); only the teacher output format differs.

\begin{table*}[t]
\caption{Per-dataset classification results for output-format ablation variants. All values are reported in (\%). \textsc{TRACE-TS} (full) uses attribution-conditioned structured DAG supervision.}
\label{tab:output_ablation_full}
\centering
\resizebox{\textwidth}{!}{%
\begin{tabular}{l|cc|cc|cc|cc|cc|cc|cc|cccc}
\toprule
& \multicolumn{2}{c|}{UCI-HAR}
& \multicolumn{2}{c|}{USC-HAD}
& \multicolumn{2}{c|}{PAMAP2}
& \multicolumn{2}{c|}{CAPTURE-24}
& \multicolumn{2}{c|}{MHEALTH}
& \multicolumn{2}{c|}{SHOAIB}
& \multicolumn{2}{c|}{OPPORTUNITY}
& \multicolumn{4}{c}{Average (7 ds)} \\
Method & Acc & F1 & Acc & F1 & Acc & F1 & Acc & F1 & Acc & F1 & Acc & F1 & Acc & F1 & Acc & F1 & $\Delta$Acc & $\Delta$F1 \\
\midrule
Label-only        & 98.30 & 98.34 & 23.76 & 19.40 & 86.42 & 80.54 & 64.60 & 61.85 & 100.00 & 100.00 & 98.77 & 98.77 & 80.93 & 74.53 & 78.97 & 76.20 &  $-$5.46 &  $-$5.04 \\
Free form output  & 70.11 & 68.34 & 11.43 &  8.75 & 70.59 & 70.74 & 51.05 & 50.07 &  87.03 &  87.37 & 89.00 & 88.93 & 67.83 & 57.44 & 63.86 & 61.66 & $-$20.57 & $-$19.58 \\
CoT reasoning     & 65.15 & 61.29 & 11.05 &  8.25 & 67.35 & 65.00 & 47.27 & 48.24 &  85.93 &  85.97 & 85.31 & 85.33 & 68.44 & 57.95 & 61.50 & 58.86 & $-$22.93 & $-$22.38 \\
\midrule
\textsc{TRACE-TS} (default) & 96.67 & 96.73 & 68.58 & 64.36 & 87.10 & 83.08 & 63.96 & 60.20 &  98.17 &  98.27 & 99.40 & 99.40 & 77.13 & 66.63 & 84.43 & 81.24 & --- & --- \\
\bottomrule
\end{tabular}%
}
\end{table*}

\subsection{Sensor Encoder and Expert Classifier Ablation: Per-Dataset Results}
\label{app:encoder_clf_ablation}

Table~\ref{tab:encoder_clf_ablation_full} reports per-dataset classification results for the sensor encoder and expert classifier ablation described in Section~\ref{sec:ablation}. All variants use Gemma~4 4B with the same cross-attention adapter configuration as \textsc{TRACE-TS}; only the upstream sensor encoder (MantisV2 $\rightarrow$ MOMENT) or the expert classifier supplying attribution targets (Attend-and-Discriminate $\rightarrow$ DeepConvLSTM) is swapped.

\begin{table*}[t]
\caption{Per-dataset classification results for the sensor-encoder and expert-classifier ablation. All values in (\%); $\Delta$ denotes change relative to the \textsc{TRACE-TS} default.}
\label{tab:encoder_clf_ablation_full}
\centering\small
\resizebox{\textwidth}{!}{%
\begin{tabular}{l|cc|cc|cc|cc|cc|cc|cc|cccc}
\toprule
 & \multicolumn{2}{c|}{UCI-HAR} & \multicolumn{2}{c|}{USC-HAD}
 & \multicolumn{2}{c|}{PAMAP2} & \multicolumn{2}{c|}{CAPTURE-24}
 & \multicolumn{2}{c|}{MHEALTH} & \multicolumn{2}{c|}{SHOAIB}
 & \multicolumn{2}{c|}{OPPORTUNITY}
 & \multicolumn{4}{c}{Average (7 ds)} \\
Method & Acc & F1 & Acc & F1 & Acc & F1 & Acc & F1 & Acc & F1 & Acc & F1 & Acc & F1 & Acc & F1 & $\Delta$Acc & $\Delta$F1 \\
\midrule
MOMENT encoder        & 76.62 & 77.51 & 55.83 & 54.72 & 74.29 & 68.31 & 58.35 & 50.63 & 72.75 & 72.64 & 96.98 & 96.98 & 19.86 &  9.26 & 64.95 & 61.43 & $-$19.48 & $-$19.81 \\
DeepConvLSTM clf.     & 92.13 & 92.33 & 14.47 &  9.23 & 85.36 & 80.36 & 63.79 & 59.98 & 90.11 & 89.70 & 97.42 & 97.40 & 73.33 & 63.49 & 73.80 & 70.35 & $-$10.63 & $-$10.89 \\
\midrule
\textsc{TRACE-TS} (default) & 96.67 & 96.73 & 68.58 & 64.36 & 87.10 & 83.08 & 63.96 & 60.20 & 98.17 & 98.27 & 99.40 & 99.40 & 77.13 & 66.63 & 84.43 & 81.24 & --- & --- \\
\bottomrule
\end{tabular}%
}
\end{table*}

\section{Attribution Analysis}

\subsection{Integrated Gradients}
\label{app:ig_heatmap}

\begin{figure}[ht]
  \centering
  \includegraphics[width=\linewidth]{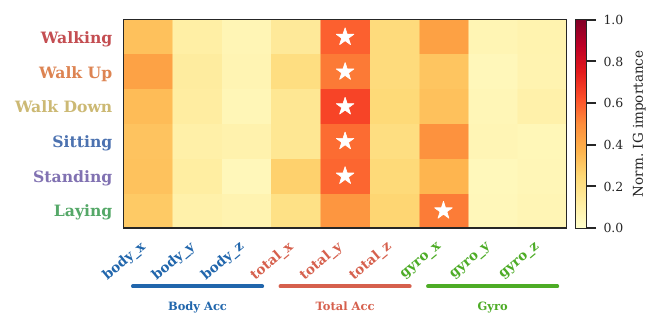}
  \Description{Heatmap of mean attribution per sensor channel for each UCI-HAR activity class, with white stars marking the dominant channel per class.}
  \caption{Per-class mean attribution per sensor channel for UCI-HAR. Rows are activity classes; columns are sensor channels grouped by modality (colored). Stars mark the dominant channel per class.}
  \label{fig:ig_heatmap_appendix}
\end{figure}

Figure~\ref{fig:ig_heatmap_appendix} shows the per-class mean IG attribution heatmap for UCI-HAR (Gemma~4 4B). Visual alignment with the cross-attention heatmap in Figure~\ref{fig:attribution}(b) confirms that the adapter attends to classifier-certified salient channels rather than arbitrary sensor tokens.

\begin{figure*}[!t]
    \centering

    \begin{subfigure}[t]{0.48\linewidth}
        \centering
        \includegraphics[width=\linewidth]{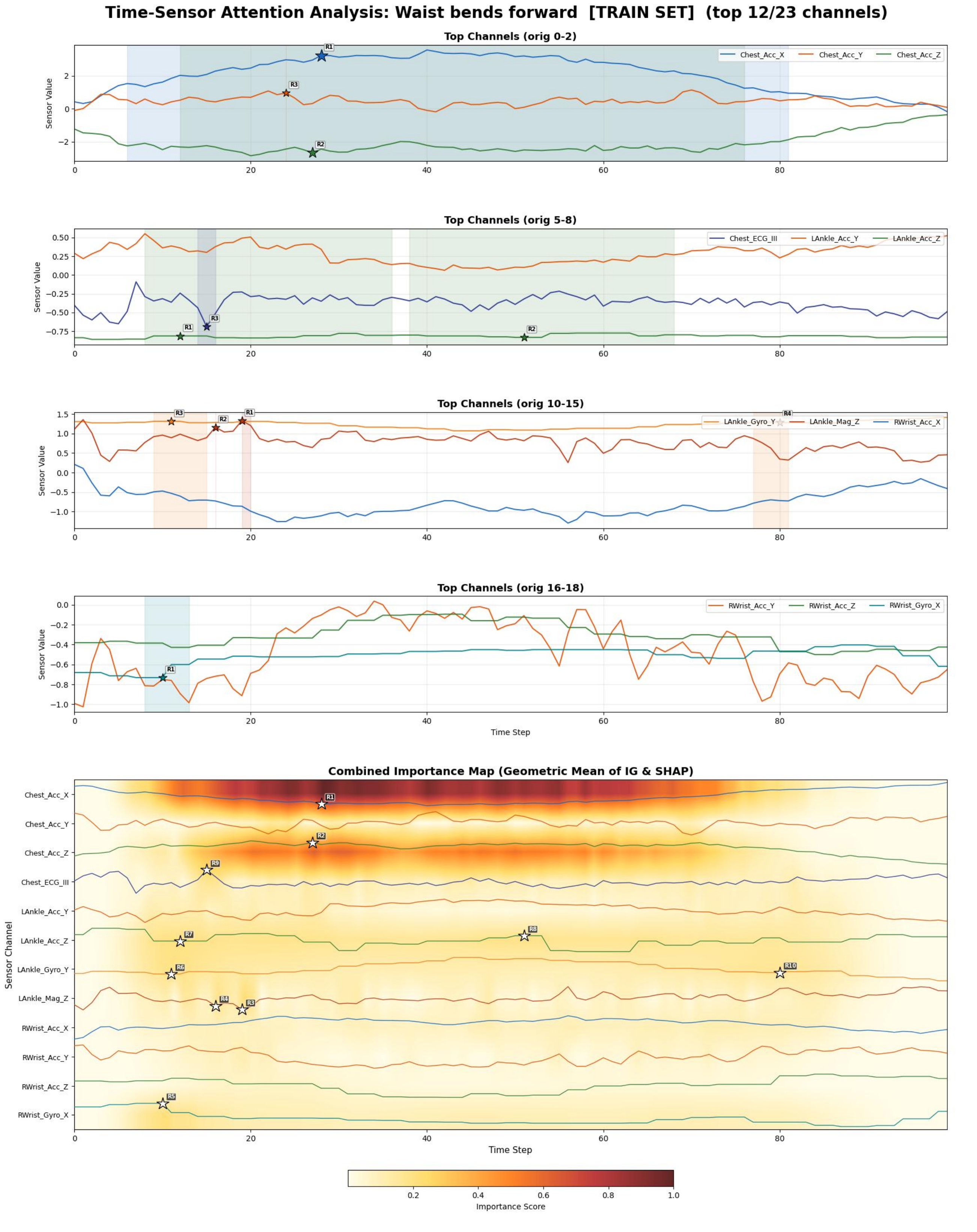}
        \Description{Attribution heatmap over sensor channels and timesteps for a waist bends forward sample, with salience concentrated on the chest accelerometer channels from the onset of trunk movement.}
        \caption{Waist bends forward from MHEALTH.}
        \label{fig:waist_bend}
    \end{subfigure}
    \hfill
    \begin{subfigure}[t]{0.48\linewidth}
        \centering
        \includegraphics[width=\linewidth]{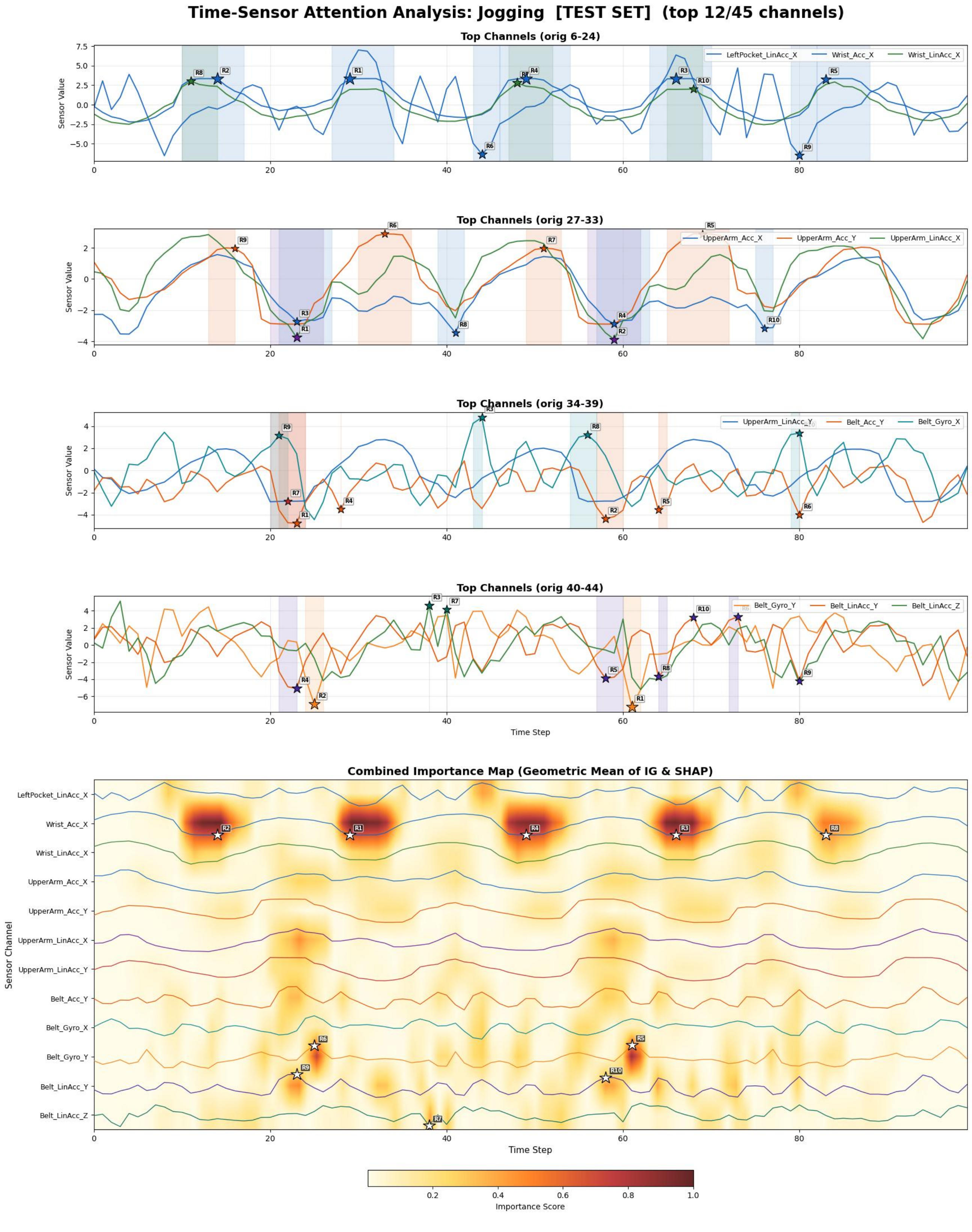}
        \Description{Attribution heatmap over sensor channels and timesteps for a jogging sample, with periodic high-salience regions along the wrist acceleration channel.}
        \caption{Jogging from SHOAIB.}
        \label{fig:jogging_shoaib}
    \end{subfigure}

    \caption{Fused IG+SHAP attribution heatmaps for two representative activity classes. Rows are sensor channels and columns are timesteps; brighter regions indicate higher salience.}
    \label{fig:qualitative_examples}
\end{figure*}
\subsection{SHAP Attribution}
\label{app:shap}

SHAP \citep{lundberg2017unified} assigns each feature $(t,c)$ its Shapley value, measuring the expected marginal contribution to the classifier output $\mathcal{C}$ over all possible feature subsets. Let $\mathcal{F}$ denote the full set of time-channel feature pairs. The SHAP attribution is:

\begin{equation}
\begin{split}
\mathbf{A}^{\text{SHAP}}_{t,c} &= \sum_{S \subseteq \mathcal{F} \setminus \{(t,c)\}} \frac{|S|!\,(|\mathcal{F}|-|S|-1)!}{|\mathcal{F}|!} \\
&\quad \times \left[\mathcal{C}(S \cup \{(t,c)\}) - \mathcal{C}(S)\right]
\end{split}
\label{eq:shap}
\end{equation}

In practice, we use KernelSHAP to approximate Eq.~\ref{eq:shap} efficiently over the sensor feature space.

\subsection{Attribution Fusion via Geometric Mean}
\label{app:fusion}

We fuse IG and SHAP through their geometric mean (Eq.~\ref{eq:attr}) rather than using either method in isolation. Unlike an arithmetic average, the geometric mean acts as a consensus operator: because it is a product, a region attains high salience only when \emph{both} methods agree it is important, and is suppressed whenever either attribution is small. This down-weights noisy regions that are quirks of a single estimator rather than genuine signal, while amplifying regions where the two methods concur. The result is a sharper, more reliable evidence map for constructing reasoning traces.

\subsection{Qualitative Attribution Analysis}
\label{app:qual_attr}

Figure~\ref{fig:qualitative_examples} shows fused attribution heatmaps for two representative classes. For \emph{jogging} (SHOAIB; Figure~\ref{fig:jogging_shoaib}), high-salience regions recur periodically along the wrist-acceleration channel, mirroring the cyclic gait of the activity. For \emph{waist bends forward} (MHEALTH; Figure~\ref{fig:waist_bend}), the chest accelerometer $x$, $y$, and $z$ channels become highly attributed precisely from the moment they begin to diverge, marking the onset of trunk movement; among the chest acceleration channels, channel $x$ carries the highest salience while $y$ has the lowest, this is consistent with forward flexion and limited lateral movement characteristic of the activity. 

\section{Qualitative Error Analysis}
\label{app:error_analysis}

To complement the aggregate ratings in Table~\ref{tab:expert-eval}, we qualitatively examined the traces evaluated in the human studies and identified two recurring patterns. The first, and dominant, error mode is confusion between fine-grained classes whose sensor signatures largely overlap. The following synthesis node was generated for an OPPORTUNITY window whose ground-truth label is \textit{Open Door}:

\begin{quote}
\small\itshape
``The data reveals a synchronized effort where the trunk twists and the arms extend forcefully. The combination of rapid acceleration in the left arm and rotation in the back, supported by leg adjustments, strongly matches the biomechanics of closing a heavy door against resistance.''
\end{quote}

The cues cited in the trace (trunk rotation, forceful arm extension, and supporting leg adjustments) characterize door manipulation in general, but none of them discriminates opening from closing: a nearly identical pushing motion can either open or close a door depending on which side of it the wearer stands, so the two classes yield almost interchangeable inertial profiles. A model whose reasoning is well grounded in the signal can therefore logically arrive at either label, and the error lies in the final class decision rather than in the reasoning chain preceding it. Such confusions between kinematically adjacent classes concentrate in fine-grained label spaces like OPPORTUNITY's door and drawer gestures, consistent with the lower recognition scores on this dataset (Appendix~\ref{app:appendix_full_results}). They also motivate scoring soundness separately from label correctness in the human evaluation (Sec.~\ref{sec:reasoning_eval}): a trace of this kind can remain logically coherent even though the predicted label is not an exact match.

The second pattern emerges from the blind pairwise comparison between attribution-grounded and no-attribution traces (Table~\ref{tab:human_ab}): ties, and the occasional preference for the no-attribution trace, concentrate on stationary activities. Tie rates are highest on USC-HAD and Capture-24, the benchmarks with the largest share of low-motion samples (Table~\ref{tab:dataset_proportions}). Table~\ref{tab:static-comparison} makes this explicit by splitting the evaluated samples into static and non-static classes, using the assignment in Table~\ref{tab:class-groups}, where the two borderline classes are resolved by their signal character: Ironing counts as static (localized arm motion around a stationary posture) and Vehicle as non-static (vehicle-induced signal dynamics dominate the window). On non-static activities the attribution-grounded trace wins the annotator majority in every sample, whereas every sample in which the no-attribution trace draws even or wins belongs to a static class. In such windows there is little for attribution to localize: all channels hold near baseline, a faithful trace reduces to reporting that no meaningful motion is present, and which region of a homogeneous window the model attends to has no bearing on that description. Both variants therefore converge on essentially the same account, leaving annotators little basis for preference. Conversely, the benefit of attribution grounding concentrates in dynamic activities, where distinctive localized patterns give the evidence records something concrete to single out.

\begin{table}[t]
\caption{Sample-level outcomes of the blind pairwise evaluation by activity type. Each sample is decided by majority over five annotator votes; \textit{Even} denotes an equal split.}
\label{tab:static-comparison}
\centering
\small
\resizebox{\columnwidth}{!}{%
\begin{tabular}{lcccc}
\toprule
Class group & $n$ & Attr.\ wins (\%) & Even (\%) & No-attr.\ wins (\%) \\
\midrule
Static     & 19 & 68.4  & 15.8 & 15.8 \\
Non-static & 31 & 100.0 & 0.0  & 0.0 \\
\bottomrule
\end{tabular}%
}
\end{table}

\begin{table}[t]
\caption{Static vs.\ non-static assignment for the activity classes present in the windows sampled for the A/B human evaluation.}
\label{tab:class-groups}
\centering
\small
\begin{tabular}{lp{0.24\columnwidth}p{0.38\columnwidth}}
\toprule
Dataset & Static & Non-static \\
\midrule
Opportunity & --- & Close Door 1, Close Door 2, Close Dishwasher, Close Drawer 2 \\
MHEALTH     & --- & Waist bends forward, Jump front \& back \\
PAMAP2      & Lying, Standing, Ironing & Walking, Running, Cycling, Ascending Stairs, Descending Stairs, Vacuum Cleaning \\
USC-HAD     & Sitting, Standing & Walking Forward, Walking Upstairs, Walking Downstairs \\
Shoaib      & --- & Walking Upstairs, Walking Downstairs \\
Capture-24  & Sleep, Sit--stand & Walking, Bicycling, Mixed, Vehicle \\
UCI-HAR     & Sitting, Laying & Walking Downstairs \\
\bottomrule
\end{tabular}
\end{table}

\onecolumn

\section{Prompt Templates}
\label{app:prompts}

Figures~\ref{fig:prompt_teacher} and~\ref{fig:prompt_student} reproduce, verbatim, the complete prompt templates used at each stage of the TRACE-TS pipeline.
Figure~\ref{fig:prompt_teacher} is the structured prompt sent to the Qwen3.5-122B-A10B teacher; angle-bracket placeholders (e.g.\ \texttt{<ACTIVITY\_LABEL>}) are substituted per sample from the IG+SHAP attribution pipeline, and the High-Importance Region and Temporal Phase blocks are emitted once per entry.
Figure~\ref{fig:prompt_student} is the fixed instruction prefix used identically during student training and inference; all sensor information is delivered through the 8 cross-attention memory tokens rather than through the text prompt.

\addvspace{\bigskipamount}

\begin{Verbatim}
You are an expert biomechanist analyzing wearable sensor data from a wearable sensor dataset. You will produce a STRUCTURED reasoning trace -- a chain of observations, inferences, and a synthesis that explains why this data corresponds to a specific activity.

## Raw Numeric Evidence (use this to ground your observations -- but paraphrase ALL numbers)

Sample ID: <SAMPLE_ID>
Predicted Activity: <ACTIVITY_LABEL>
Confidence: <CONFIDENCE>
Attribution threshold (global p90): <P90_THRESHOLD>

### High-Importance Regions (sorted by importance):
[one block below per region, i = 1 .. <N_REGIONS> (up to 10)]
<i>. Sensor: <SENSOR_NAME> (original: <RAW_SENSOR_NAME>), Timesteps <T_START>-<T_END> (length <LENGTH>)
   - Mean Importance: <MEAN_IMPORTANCE>
   - Max Importance:  <MAX_IMPORTANCE>
   - Peak Timestep:   <T_PEAK>
   - Temporal Region: <TEMPORAL_LABEL>
   - Confidence Tier: <CONF_TIER>

### Temporal Phase Analysis:
[one line below per phase]
Phase <P>: Average Importance <PHASE_IMPORTANCE>

## YOUR TASK: Generate a Structured Reasoning Trace

You must output EXACTLY this format. Every section header must appear EXACTLY as shown (including the brackets and pipe characters). Do NOT add any other headers, bullet points, or markdown.

### FORMAT SPECIFICATION:

**STEP 1 -- OBSERVATIONS** (one per important sensor region)
Generate exactly <N_REGIONS> observations, one for each high-importance region listed above. Each observation describes what a specific sensor is doing in a specific temporal window.

[OBSERVATION | id: O1]
sensor: <sensor_name from the evidence, using normalized lowercase format>
temporal: <EXACTLY one of: early | early_to_mid | mid | mid_to_late | late | full_window>
pattern: <faithfully describe only what the evidence above shows this sensor doing -- peaks, oscillations, stability, sustained elevation, sudden transitions, etc. Do NOT invent patterns not present in the evidence. Use natural language, NO numbers.>
confidence: <high, moderate, or low -- use the confidence tier from the evidence>

**STEP 2 -- INFERENCES** (combine observations into biomechanical interpretations)
Generate 2-4 inferences. Each inference MUST cite which observations it builds on using their IDs (O1, O2, etc.).

[INFERENCE | id: I1]
based_on: O1, O3
inference: <what the cited observations actually imply biomechanically -- heel strikes, push-off phases, postural stability, arm swing, weight transfer, etc. Must follow directly from the observations cited in based_on. Do not introduce evidence not present in those observations.>
confidence: <high, moderate, or low>

**STEP 3 -- SYNTHESIS** (combine inferences into final explanation)
One synthesis block that ties everything together. Must cite which inferences it builds on.

[SYNTHESIS]
based_on: I1, I2
<A coherent 50-80 word paragraph explaining why the overall sensor profile indicates this activity. Reference the key biomechanical mechanisms. NO numbers -- speak in terms of body mechanics.>

**STEP 4 -- ACTIVITY**

[ACTIVITY]: <the activity name, exactly as given: <ACTIVITY_LABEL>>

### CRITICAL RULES:
1. sensor names MUST use one of these exact values: body_acc_x, body_acc_y, body_acc_z, gyro_x, gyro_y, gyro_z, total_acc_x, total_acc_y, total_acc_z
2. temporal MUST be EXACTLY one of: early | early_to_mid | mid | mid_to_late | late | full_window
   Replace any free-form phrase (e.g. "beginning", "around the middle", "toward the end") with the closest canonical label.
3. pattern (OBSERVATION) MUST faithfully describe only what the evidence shows. DO NOT INVENT SENSOR BEHAVIOURS WHICH ARE NOT PRESENT.
4. inference MUST follow directly from the observations cited in its based_on field -- do not introduce biomechanical claims that are not grounded in those specific observations.
5. [SYNTHESIS] MUST have a based_on field citing inference IDs (I1, I2, etc.)
6. Do NOT use any numbers, percentages, or timestep values in any text
7. Do NOT add any markdown formatting, code fences, headers, or bullet points outside the specified format
8. Observation IDs must be sequential: O1, O2, O3, ... O<N_REGIONS>
9. Inference IDs must be sequential starting from I1
10. Output ONLY the structured trace -- no preamble, no explanation, no commentary
12. All output shall be in English only -- no Chinese or any other language

### PARAPHRASING RULES:
Instead of numbers, use natural language:
- Timesteps -> "early portion", "midway through", "toward the end"
- Sensor values -> "sharp peak", "rapid oscillation", "sustained elevation"
- Importance -> "strong evidence", "a secondary signal"

### LANGUAGE RULES -- READ CAREFULLY:
Your pattern descriptions MUST be plain, direct, and mechanically grounded.
FORBIDDEN language styles -- outputs containing these will be rejected:

- Dramatic and abstract language. this is not a language exam, you need to write everything in a manner that a layman can understand.
- Padding phrases: "occurring within the initial segment", "distributed across the central portion of the timeline" are FORBIDDEN
- Any phrase that sounds like it is trying to sound impressive rather than understandable, KEEP IT SIMPLE,

REQUIRED language style:
- Describe what the signal IS DOING in the simplest possible terms
- Maximum 12 words per pattern description
- If nothing significant is happening, SAY THAT PLAINLY:
  BAD:  "subtle undulations indicating gentle sway rather than violent motion"
  GOOD: "signal holds near baseline with no significant change"
  BAD:  "isolated single point deviation indicating a negligible side-to-side shift"
  GOOD: "near-flat signal with one small spike, otherwise no movement"


PHYSIOLOGICAL SENSORS (HR, EDA, temperature only)
Before treating an anomaly as real, judge whether it's noise or artifact. For example, wrist HR, temperature sensors are prone to noise during movement due to loss of contact with body.

Keep noisy sections if a reasonable cause exists, like heart rate elevated by orders of magnitude in walking. This is CLEARLY due to sensor losing contact with body intermittently.
IF YOU KEEP THE NOISE IN THE REASON, MENTION THAT IT IS NOISE, AND THE REASON (loss of contact etc) ALONG WITH IT.
Discard if no plausible cause. Never surface noise as a physiological event.

### STATIC ACTIVITY RULE (sitting, standing, lying down):
If the predicted activity is a static or low-motion activity (Standing, Sitting, Sitting Down,
Standing Up, Laying), your observations MUST reflect that directly.
- Do NOT dress up flat signals with dramatic vocabulary
- If a sensor holds near a constant value, write: "holds near [high/low/baseline] throughout"
- If there are only tiny fluctuations, write: "small fluctuations around a stable value, no clear movement"

### BAD vs GOOD PATTERN EXAMPLES (study these carefully):

BAD: "distinct sharp fluctuations with extreme intensity peaks occurring within the initial segment"
GOOD: "sharp peak then drops back to baseline"

BAD: "intermittent bursts of energy signaling brief shifts in vertical load distribution"
GOOD: "small oscillations around a stable baseline"

BAD: "localized surges in horizontal force corresponding to minor adjustments"
GOOD: "minor variations, no clear directional movement"


### SELF-CHECK before writing each pattern:
Ask yourself: "Is this phrase something that the layperson can understand? or does it sound like creative writing, or something with a lot of fancy words for no reason?" If creative writing -> rewrite it.

### EXAMPLE OUTPUT:

[OBSERVATION | id: O1]
sensor: total_acc_y
temporal: early
pattern: sharp peak then stabilizes near a constant value
confidence: high

[OBSERVATION | id: O2]
sensor: total_acc_y
temporal: late
pattern: small fluctuations around a stable value, no clear movement
confidence: high

[OBSERVATION | id: O3]
sensor: total_acc_y
temporal: mid
pattern: holds near constant, minimal variance throughout
confidence: high

[OBSERVATION | id: O4]
sensor: body_acc_x
temporal: mid
pattern: stays near zero, no meaningful lateral movement
confidence: moderate

[OBSERVATION | id: O5]
sensor: gyro_z
temporal: full_window
pattern: near-zero throughout, confirming no rotation
confidence: moderate

[INFERENCE | id: I1]
based_on: O1, O2, O3
inference: vertical signal holds steady after initial settling, consistent with remaining upright and still
confidence: high

[INFERENCE | id: I2]
based_on: O4, O5
inference: no lateral movement or rotation detected, ruling out walking or any locomotion
confidence: high

[SYNTHESIS]
based_on: I1, I2
The vertical axis holds near a constant value with no rhythmic pattern. Lateral and rotational signals stay near zero throughout. There is no stepping, turning, or directional movement detectable. This is consistent with the subject standing still.

[ACTIVITY]: Standing

Now generate the structured reasoning trace for **<ACTIVITY_LABEL>**:
\end{Verbatim}
\rawvspace{-3pt}
\noindent\begin{minipage}{\linewidth}
\captionsetup{hypcap=false}
\Description{Verbatim text of the structured teacher prompt, including evidence records, format specification, critical rules, paraphrasing and language rules, and an example output.}
\captionof{figure}{Full, verbatim teacher reasoning-generation prompt for UCI-HAR (Qwen3.5-122B-A10B); the output vocabulary is correspondingly its 9 sensor channels and 6 temporal terms. The rule list skips index ``11'' in the source prompt and is reproduced as-is.}
\label{fig:prompt_teacher}
\end{minipage}

\addvspace{\bigskipamount}

\begin{Verbatim}
User: Analyze the sensor embeddings, generate reasoning and explain the activity:
Assistant: <student-generated structured reasoning trace + activity label>
\end{Verbatim}
\rawvspace{-3pt}
\noindent\begin{minipage}{\linewidth}
\captionsetup{hypcap=false}
\Description{Verbatim text of the two-line student prompt: a user instruction to analyze the sensor embeddings and the assistant's generated reasoning trace with activity label.}
\captionof{figure}{Full, verbatim student-model prompt, identical at training and inference.}
\label{fig:prompt_student}
\end{minipage}

\section{Usage of AI Tools}
AI based tools were used for limited low-level support tasks, including language refinement, grammar correction, formatting assistance, and minor debugging or code-review related activities. These tools were not used to generate scientific claims, interpret results, design experiments, or formulate the methodology. All core research contributions including problem formulation, experimental design, methodological development, implementation, analysis, interpretation of findings, and scientific conclusions were independently performed, verified, and validated by the authors.

\section{Ethical Considerations}
\label{sec:ethics}

\paragraph{Use of existing artifacts and consistency with intended use.}
All seven datasets used in this work (UCI-HAR, USC-HAD, PAMAP2, CAPTURE-24, MHEALTH, SHOAIB, and OPPORTUNITY) are publicly released human activity recognition benchmarks intended for research on wearable sensing and activity recognition. Our use, namely training and evaluating activity-recognition and reasoning models, is consistent with this intended research use, and we credit each dataset through citation. The pretrained models we build on (Gemma~4, Llama~3.2/3.3, Qwen~3.5, MantisV2, and Chronos-2) and the software libraries we rely on are likewise used within the research scope permitted by their licenses. Full per-artifact license names, sources, and terms of use are provided in Appendix~\ref{sec:appendix_licensing}.

\paragraph{Privacy and anonymization.}
The datasets contain de-identified inertial and physiological sensor recordings collected from consenting volunteer participants in the original studies, and are distributed without personally identifiable information; participants are referenced only by anonymous subject identifiers. We perform no re-identification and infer no protected attributes (e.g., gender, sexual orientation, or health status). All processing operates at the level of motion-sensor windows and activity labels.

\paragraph{Intended use of created artifacts and access conditions.}
We release our code and the derived structured reasoning-trace annotations strictly as a research prototype, under terms compatible with the original datasets' research-oriented access conditions (Appendix~\ref{sec:appendix_licensing}). Because the source data are provided for research, the derived traces and trained adapters should not be deployed as anything other than a research artifact, and in particular not as a clinical, diagnostic, or commercial activity-monitoring system without separate validation, participant consent, and regulatory review.

\paragraph{Potential risks.}
\textsc{TRACE-TS} produces human-readable reasoning over body-motion signals. While this improves transparency, generated reasoning traces are bounded by the teacher LLM and the attribution pipeline (Sec.~\ref{sec:method}) and may contain errors; they should not be treated as verified clinical or behavioral assessments. As with other wearable activity models, deployment in real-world monitoring settings could raise surveillance and privacy concerns for monitored or vulnerable populations, and should be governed by informed consent and appropriate data-protection safeguards.

\end{document}